%% file: main.tex
\documentclass[10pt,twocolumn,letterpaper]{article}

\usepackage[pagenumbers]{cvpr} 

\usepackage{comment}
\usepackage{soul}
\usepackage{multirow}
\usepackage{multicol}
\usepackage{float}
\usepackage{cuted}
\usepackage{lipsum}
\usepackage{marvosym}
\usepackage{dsfont}

\usepackage[dvipsnames]{xcolor}
\usepackage{colortbl}

\definecolor{tx_color}{RGB}{200,42,40}
\definecolor{delete_red}{RGB}{255,72,32}
\definecolor{fweS_color}{RGB}{255,0,0}
\definecolor{weBc_color}{RGB}{0,255,0}
\definecolor{jaza_color}{RGB}{0,0,255}

\newcommand{\OURS}{FastGS\xspace}%
\newcommand{\VCD}{VCD}
\newcommand{\VCP}{VCP}

\definecolor{tabfirst}{rgb}{1, 0.6, 0.6} 
\definecolor{tabsecond}{rgb}{1, 0.8, 0.5} 
\definecolor{tabthird}{rgb}{1, 1, 0.6} 

\definecolor{tabfirst}{rgb}{1, 0.6, 0.6} 
\definecolor{tabsecond}{rgb}{1, 0.8, 0.5} 
\definecolor{tabthird}{rgb}{1, 1, 0.6} 

\definecolor{lightred}{rgb}{1, 0.6, 0.6}
\definecolor{lightorange}{rgb}{1, 0.8, 0.5}
\definecolor{lightyellow}{rgb}{1, 1, 0.6}

\usepackage{amsmath}

\definecolor{cvprblue}{rgb}{0.21,0.49,0.74}
\usepackage[pagebackref,breaklinks,colorlinks,allcolors=cvprblue]{hyperref}
\usepackage{xcolor}

\def\paperID{***} 
\def\confName{CVPR}
\def\confYear{2025}

\title{\OURS{}: Training 3D Gaussian Splatting in 100 Seconds}

\author{
Shiwei Ren\textsuperscript{*}~~~~~Tianci Wen\textsuperscript{*}~~~~~Yongchun Fang\textsuperscript{\Cross}~~~~~Biao Lu
\\  NanKai University\\
{\tt\small renshiwei, wentc, lubiao@mail.nankai.edu.cn, fangyc@nankai.edu.cn}\\
[0.75em]
\url{https://fastgs.github.io}
}

\begin{document}
\input{sec/0_abstract}
\newcommand\blfootnote[1]{%
  \begingroup
  \renewcommand\thefootnote{}\footnote{#1}%
  \addtocounter{footnote}{-1}%
  \endgroup
}
\blfootnote{$^{*}$Equal contribution.}
\blfootnote{$^\text{\Cross}$Corresponding author.}
\input{sec/1_intro}
\input{sec/2_related}
\input{sec/3_background}
\input{sec/4_method}
\input{sec/5_experiment}
\input{sec/6_conclusion}

{
    \small
    \bibliographystyle{ieeenat_fullname}
    \bibliography{main}
}

\input{sec/X_suppl}

\end{document}

%% file: sec/0_abstract.tex
\input{tables_and_figures/image1_teaser}
\begin{abstract}
The dominant 3D Gaussian splatting (3DGS) acceleration methods fail to properly regulate the number of Gaussians during training, causing redundant computational time overhead. In this paper, we propose \OURS{}, a novel, simple, and general acceleration framework that fully considers the importance of each Gaussian based on multi-view consistency, efficiently solving the trade-off between training time and rendering quality. We innovatively design a densification and pruning strategy based on multi-view consistency, dispensing with the budgeting mechanism. Extensive experiments on Mip-NeRF 360, Tanks \& Temples, and Deep Blending datasets demonstrate that our method significantly outperforms the state-of-the-art methods in training speed, achieving a 3.29$\times$ training acceleration and comparable rendering quality compared with DashGaussian on the Mip-NeRF 360 dataset and a 15.45$\times$ acceleration compared with vanilla 3DGS on the Deep Blending dataset. We demonstrate that \OURS{} exhibits strong generality, delivering 2-6$\times$ training acceleration across various tasks, including dynamic scene reconstruction, surface reconstruction, sparse-view reconstruction, large-scale reconstruction, and simultaneous localization and mapping.
\end{abstract}

%% file: tables_and_figures/image1_teaser.tex
\begin{figure}[t]
    \vspace{-6mm}
    \twocolumn[{
        \renewcommand\twocolumn[1][]{#1}
        \maketitle
        \centering
        \setlength\tabcolsep{0pt}
        \includegraphics[width=\linewidth]{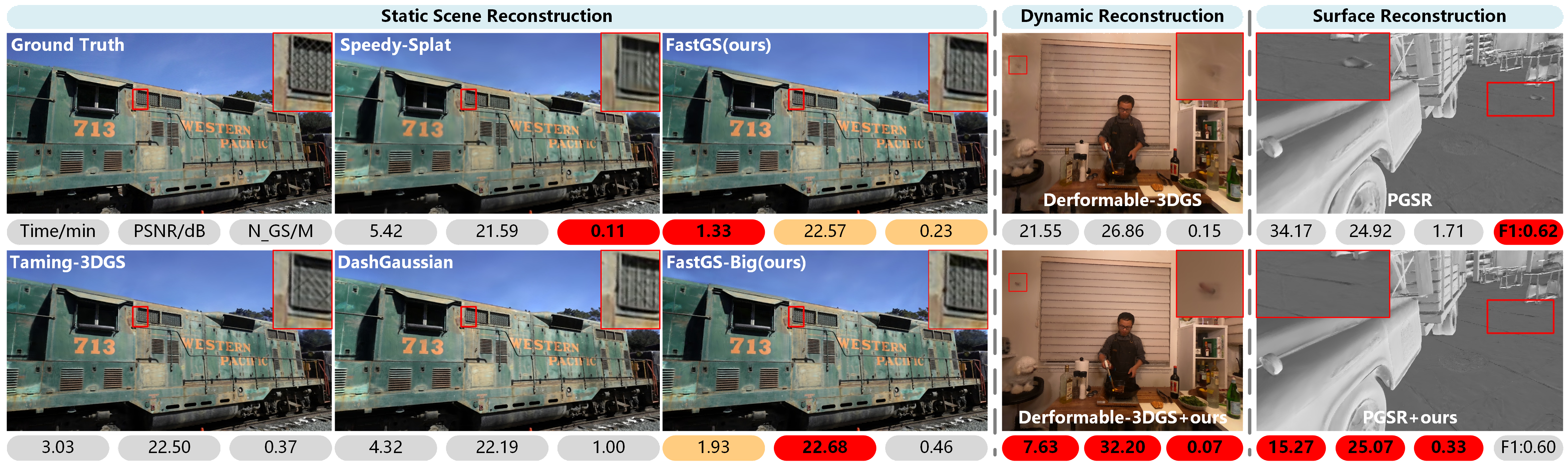}
        \vspace{-6mm}
        \caption{
            We propose \OURS{}, a general acceleration framework for 3D Gaussian Splatting (3DGS) that significantly reduces training time without sacrificing rendering quality. In static scenes, our method completes training on the Tanks \& Temples \textit{train} scene within 100 seconds. Furthermore, our method achieves 2.82× and 2.24× faster training for dynamic and surface reconstruction, respectively. 
        }
        \label{fig:teaser}
        \vspace{2mm}
    }]
\end{figure}

%% file: sec/1_intro.tex
\section{Introduction}
\label{sec:intro}
Novel view synthesis (NVS) is a fundamental problem in computer vision and graphics, with broad applications in augmented reality~\cite{zhou2018stereo}, virtual reality~\cite{xu2023vr}, and autonomous driving~\cite{wu2023mars}. Neural Radiance Field (NeRF)~\cite{mildenhall2021nerf} methods model scenes as continuous volumetric functions and render photorealistic views, but require hours of training per scene. Recently, 3D Gaussian Splatting (3DGS)~\cite{kerbl3Dgaussians} has achieved rendering quality comparable to NeRF while offering significantly faster training and rendering speed. It models 3D scenes via explicit Gaussian primitives and employs a tile-based rasterizer. Benefiting from its efficiency, 3DGS has been successfully applied to a wide range of tasks, including dynamic scene reconstruction, surface reconstruction, and simultaneous localization and mapping (SLAM). However, a major bottleneck in its current practical application is the extended training time, which often requires tens of minutes per scene, hindering user-friendly deployment.

A detailed analysis of the vanilla 3DGS~\cite{kerbl3Dgaussians} training pipeline reveals two primary limitations: (1) its adaptive
density control (ADC) of Gaussians often introduces numerous redundant Gaussians and (2) inefficiencies in the rendering pipeline. While recent works~\cite{mallick2024taming,hanson2025speedy,feng2025flashgs,gui2024balanced} have significantly optimized the rendering pipeline, ADC remains a major area for improvement.

ADC in vanilla 3DGS~\cite{kerbl3Dgaussians} comprises two main components. The first is Gaussian densification, which clones or splits a Gaussian based on its positional gradient. The second is Gaussian pruning, which removes Gaussians with low opacity or oversized scales. Existing 3DGS acceleration methods~\cite{fang2024mini,girish2024eagles,papantonakis2024reducing,wang2024adr,deng2024efficient,mallick2024taming,hanson2025speedy,chen2025dashgaussian,kim2024color,hanson2025pup} have introduced improvements to ADC. One direction of improvement focuses on designing mechanisms to constrain Gaussian densification, aiming to minimize the growth of redundant Gaussians. For example, Taming-3DGS~\cite{mallick2024taming} employs a budget-constrained optimization to control Gaussian growth. Similarly, DashGaussian~\cite{chen2025dashgaussian} leverages an adaptive Gaussian primitive budgeting method to maintain continuous densification throughout training. The other direction involves refining the pruning strategy to accelerate training by deleting a greater number of Gaussians. For example, Speedy-Splat~\cite{hanson2025speedy} applies a soft pruning strategy during densification and a hard pruning strategy afterward.

One major drawback of these methods is the limited effectiveness of their densification and pruning strategies, which fail to maintain rendering quality while avoiding excessive Gaussian redundancy, resulting in an inefficient representation, as illustrated in \cref{fig:gaussian_count}. This indicates that their Gaussian control strategies are suboptimal. Based on our observations, some densification and pruning methods~\cite{kim2024color,chen2025dashgaussian,girish2024eagles,salman2024elmgs} do not leverage multi-view consistency, while others~\cite{mallick2024taming,hanson2025speedy,hanson2025pup} exploit it suboptimally. Specifically, they enforce multi-view consistency merely through Gaussian-associated scores, which we argue is insufficient. It may lead to the excessive growth of redundant Gaussians that provide only marginal improvements to the rendering quality from a few viewpoints while contributing little to others. On one hand, for certain densification methods such as Taming-3DGS~\cite{mallick2024taming}, Gaussian importance is considered across views. However, it fully relies on Gaussian-associated scores rather than their actual contribution to rendering quality, resulting in weak multi-view constraints and leading to redundancy. Moreover, it lacks a dedicated redesign of the pruning strategy. On the other hand, for some pruning methods, such as Speedy-Splat~\cite{hanson2025speedy}, multi-view information is also considered, but it uses the gradients of Gaussian rather than evaluating each Gaussian’s contribution to multi-view rendering quality. This indirect enforcement of multi-view consistency leads to significant degradation in rendering quality.

To address the above issues, we propose \OURS{}, a new, simple, and general  3DGS acceleration framework, capable of training a scene in around 100 seconds while maintaining comparable rendering quality, as shown in \cref{fig:teaser}. In fact, nearly every Gaussian primitive participates in rendering the same region across multiple viewpoints. Our insight is similar to the concept behind bundle adjustment in traditional 3D reconstruction, where each 3D Gaussian should maintain multi-view consistency. This implies that the 3D Gaussian should enhance rendering quality across multiple views of the same region. Therefore, we introduce a multi-view consistent densification (VCD) strategy, which uses a multi-view reconstruction quality importance score to evaluate whether a Gaussian contributes beneficially to the improvement of multi-view rendering quality. Based on the same idea, we propose a multi-view consistent pruning (VCP) strategy, which removes redundant Gaussians that are useless to multi-view rendering quality. Notably, because VCD and VCP accurately identify which Gaussians need to be densified or pruned, our method does not require a budget mechanism, making it easily applicable to other tasks. To summarize our contributions,

\input{tables_and_figures/image_gaussian_count}

\begin{itemize}
    \item[1.]
\textbf{New, simple, and general framework} for  3DGS acceleration that can train a scene in around 100 seconds while achieving comparable rendering quality.
    \item[2.]
\textbf{Efficient densification and pruning strategy} strictly controlling the addition and removal of each Gaussian based on its contribution to multi-view reconstruction quality, greatly accelerating the training process.
    \item[3.]
    \textbf{General and state-of-the-art performance} across various tasks. Our method outperforms state-of-the-art (SOTA) methods in training speed while maintaining comparable rendering quality on static scenes. It generalizes well to dynamic scene reconstruction, surface reconstruction, sparse-view reconstruction, large-scale reconstruction, and SLAM. 
\end{itemize}

%% file: tables_and_figures/image_gaussian_count.tex
\begin{figure}[t]
    \centering
    \captionsetup{singlelinecheck=false}
    \includegraphics[width=\linewidth]{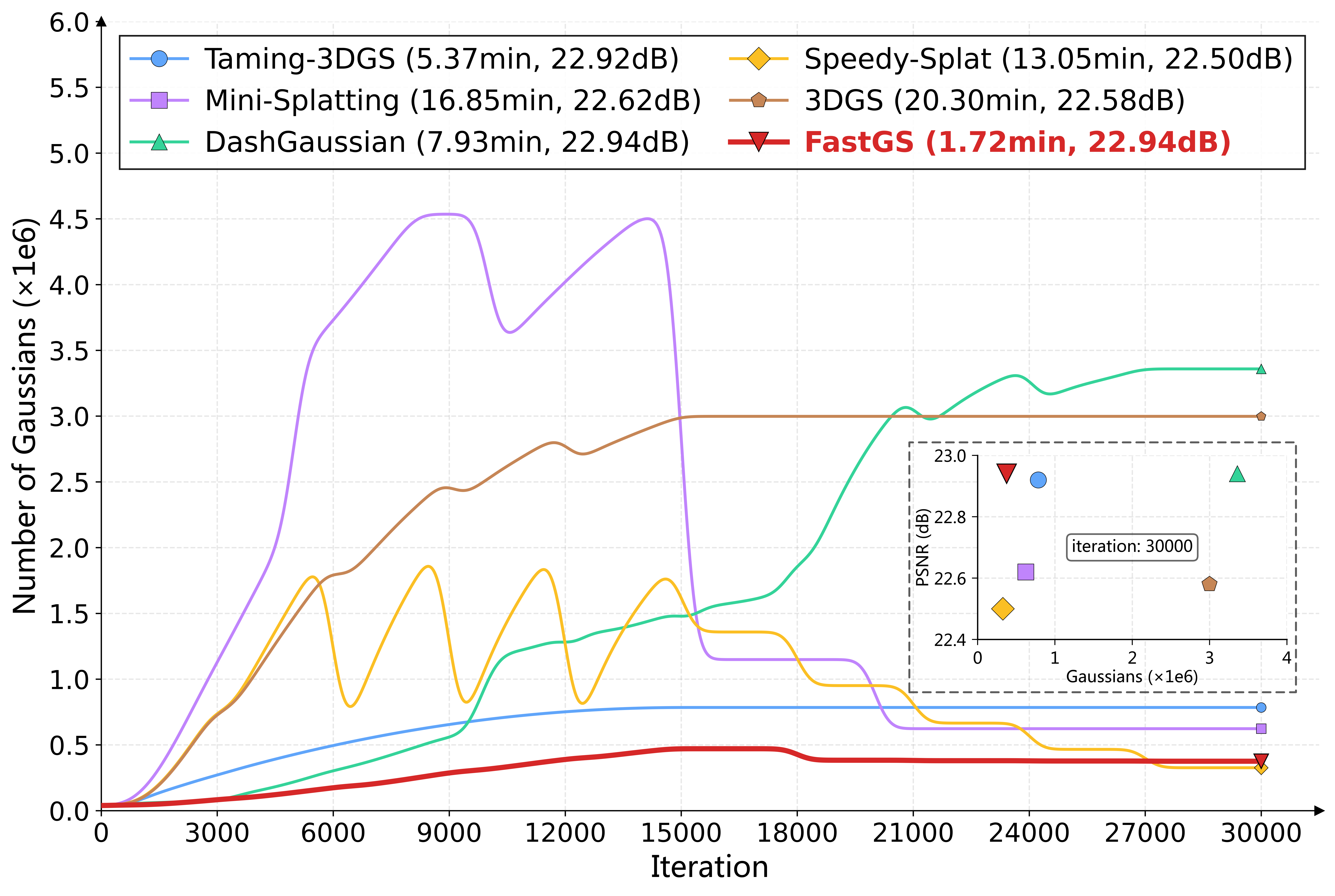}
    \caption{ 
         \textbf{Gaussian count over training iterations.}
          Benefiting from the efficient VCD and VCP strategies, \OURS{} keeps the number of Gaussians consistently low throughout the entire training process on the \textit{treehill} scene of Mip-NeRF 360~\cite{barron2022mip}. 
    }
    \label{fig:gaussian_count}
\end{figure}

%% file: sec/2_related.tex
\section{Related Work}
\label{sec:related}
In this section, we review related works that
focus on accelerating both the training and inference of 3DGS~\cite{kerbl3Dgaussians}.

\noindent\textbf{Gaussian Densification.} A key bottleneck of 3DGS is the excessive Gaussians that significantly slow down the training process. Recent works~\cite{kim2024color,rota2024revising,mallick2024taming,chen2025dashgaussian, deng2024efficient} attempt to address this issue by refining the densification strategy to control the primitive count. Specifically, Taming-3DGS~\cite{mallick2024taming} employs a budgeting mechanism to control Gaussian growth, primarily computing importance scores based on Gaussian-associated properties. DashGaussian~\cite{chen2025dashgaussian} introduces a resolution-guided primitive scheduler that progressively reconstructs the scene throughout the entire training process. Nevertheless, they still require millions of Gaussians to maintain rendering quality, resulting in heavy optimization overhead.

\noindent\textbf{Gaussian Pruning.} Besides modifying densification, some methods~\cite{ali2024trimming,salman2024elmgs,hanson2025speedy,hanson2025pup,girish2024eagles,fang2024mini,papantonakis2024reducing} achieve acceleration by designing pruning strategies to remove a large number of Gaussians. Some of them design importance scores according to Gaussian-associated properties to guide pruning, discarding less critical Gaussians~\cite{ali2024trimming,girish2024eagles,salman2024elmgs}. Mini-Splatting~\cite{fang2024mini} removes a large number of Gaussians through a simplification strategy based on intersection preserving and sampling. PUP 3DGS~\cite{hanson2025pup} and Speedy-Splat~\cite{hanson2025speedy} remove Gaussians by computing a Gaussian-associated Hessian approximation across all training views. Nonetheless, some of them fail to fully remove redundant Gaussians, while others remove many Gaussians at the cost of significantly degraded rendering quality.

\noindent\textbf{Other Methods.} Some works focus on optimizing 3DGS rasterization or optimization strategies. Taming-3DGS~\cite{mallick2024taming} replaces per-pixel with per-splat parallel backpropagation, which significantly speeds up the
 optimization process and serves as a strong baseline for
 subsequent research. StopThePop~\cite{radl2024stopthepop}, FlashGS~\cite{feng2025flashgs}, and Speedy-Splat~\cite{hanson2025speedy} use precise tile intersection to reduce Gaussian-tile pairs and accelerate rasterization. 3DGS-LM~\cite{hollein20243dgs} replaces Adam with Levenberg-Marquardt for faster convergence, and 3DGS²~\cite{lan20253dgs2} achieves near second-order convergence via prioritized per-kernel updates.

%% file: sec/3_background.tex
\section{Background: 3D Gaussian Splatting}
\label{sec:prelim}
3DGS models a scene as an explicit point-based representation composed of a set of anisotropic 3D Gaussians:
\begin{equation}
\left\{ \mathcal{G}_i(\mathbf{x}) = \exp\!\left(-\tfrac{1}{2}(\mathbf{x}-\boldsymbol{\mu}_i)^{\top}\mathbf{\Sigma }_{i_{3D}}^{-1}(\mathbf{x}-\boldsymbol{\mu}_i)\right) \right\}_{i=1}^N .
\end{equation}
To construct this representation, a sparse point cloud obtained from SfM is used to initialize the positions of Gaussian primitives. 
Each primitive $\mathcal{G}_i$ is parameterized by mean $\boldsymbol{\mu}_i \in \mathbb{R}^3$, rotation $r_i \in \mathbb{R}^4$, scale $s_i \in \mathbb{R}^3$, opacity $\sigma_i \in \mathbb{R}$, and color coefficients $\mathbf{c}_i \in \mathbb{R}^{16\times 3}$ represented in view-dependent spherical harmonics (SH).
The rotation and scale together define the covariance matrix as
\begin{equation}
\mathbf{\Sigma}_{i_{3D}} = R_i S_i S_i^{\top} R_i^{\top}.
\end{equation}

To render from a given camera viewpoint, all 3D Gaussians need to be projected into the 2D image plane. 
Given a viewing transformation $W$, the covariance matrix in camera coordinates is computed as
\begin{equation}
\boldsymbol{\Sigma}'_{i_{3D}} = JW \, \boldsymbol{\Sigma}_{i_{3D}} \, W^\top J^\top,
\end{equation}
where $J$ denotes the Jacobian of the affine approximation of the projective transformation. The projected 3D Gaussian is then approximated as a 2D elliptical Gaussian on the image plane, with covariance $\boldsymbol{\Sigma}_{i_{2D}}$ obtained by marginalizing $\boldsymbol{\Sigma}'_{i_{3D}}$ along the viewing direction. 
Each 2D Gaussian contributes to pixels within its footprint using $\alpha$-blending: given Gaussians $\{\mathcal{G}_i\}_{i=1}^N$ sorted by depth, the accumulated color of pixel $p$ is computed as
\begin{equation}
C(p)=\sum_{i\in N}{c_i\alpha _i\prod_{j=1}^{i-1}{\left( 1-\alpha _j \right)}}, \quad 
\alpha_i=\sigma _i\mathcal{G} _{i}^{\prime}(p),
\end{equation}
where
\begin{equation}
\mathcal{G}'_i(p) = \exp\Big(-\tfrac{1}{2}(p-\boldsymbol{\mu}_{i_{2D}})^{\top} \boldsymbol{\Sigma}_{i_{2D}}^{-1} (p-\boldsymbol{\mu}_{i_{2D}}) \Big),
\end{equation}
with $\boldsymbol{\mu}_{i_{2D}}$ and $\boldsymbol{\Sigma}_{i_{2D}}$ denoting the mean and covariance of the projected 2D Gaussian, respectively.

%% file: sec/4_method.tex
\input{tables_and_figures/image2_framework}
\section{\OURS{}}
\label{sec:method}
\subsection{Overview}
\label{sec:method:overview}
The framework of our method is shown in \cref{fig:pipeline}. We initialize 3D Gaussians using SfM point clouds and train the 3DGS model on multi-view images. The addition and removal of 3D Gaussians are controlled via the proposed multi-view consistent densification and pruning strategies. As illustrated in Fig.~\ref{fig:pipeline:b} and Fig.~\ref{fig:pipeline:c}, Taming-3DGS~\cite{mallick2024taming} and Speedy-Splat~\cite{hanson2025speedy} also estimate importance scores from multiple views, for densification and pruning, respectively. However, both rely on Gaussian-associated scores to control the number of Gaussians. This suboptimal use of multi-view information results in redundancy for Taming-3DGS~\cite{mallick2024taming} and degraded rendering quality for Speedy-Splat~\cite{hanson2025speedy}. In contrast, as illustrated in Fig.~\ref{fig:pipeline:a}, our method evaluates the importance of each Gaussian based on multi-view reconstruction quality, rather than gaussian-associated properties. Furthermore, our method leverages multi-view consistency constraints to effectively guide both densification and pruning, which will be detailed in Sections Sec.~\ref{sec:method:densification} and Sec.~\ref{sec:method:pruning}. To further improve rasterization efficiency, Sec.~\ref{sec:method:focusedbox} introduces the compact box (CB) we use, which is adapted from the precise tile-intersection strategy proposed in Speedy-Splat~\cite{hanson2025speedy}.

\subsection{Multi-view Consistent Densification}
\label{sec:method:densification}
The vanilla 3DGS~\cite{kerbl3Dgaussians} densifies Gaussians solely based on the gradient magnitude in the image space, which leads to a large number of redundant Gaussians. Other densification methods~\cite{chen2025dashgaussian,ye2024absgs,deng2024efficient} also generate millions of Gaussians, leading to inefficiency. We argue that the redundancy arises because these methods fail to rigorously determine from multiple views whether a Gaussian needs densification.
As illustrated in Fig.~\ref{fig:pipeline:b}, Taming-3DGS~\cite{mallick2024taming} considers multi-view consistency during densification. However, it primarily computes the score based on Gaussian-associated properties (e.g., opacity, scale, depth, and gradient), making it difficult to enforce strict multi-view consistency for a Gaussian. This also leads to redundancy, as visualized on the left of Fig.~\ref{fig:pipeline:b}. Moreover, the computation of its score is complex and relatively inefficient. To address these issues, we propose a new, simple densification strategy \VCD{} based on multi-view consistency. As illustrated in Fig.~\ref{fig:pipeline:a}, it computes the average number of high-error pixels in each Gaussian’s 2D footprint across sampled views, where high-error pixels are identified solely from the per-pixel L1 loss between the ground truth and the rendering. As shown on the left of Fig.~\ref{fig:pipeline:a}, VCD achieves comparable rendering quality with fewer Gaussians, thereby greatly avoiding redundancy.  We then detail how \VCD{} is implemented.

Given $K$ camera views $V=\{v^j\}_{j=1}^K$, randomly sampled from the training views, together with their corresponding ground-truth images $G=\{g^j\}_{j=1}^K$ and rendered images $R=\{r^j\}_{j=1}^K$. For each view $v^j$, we compute the error between the rendered color $r^{j}_{u,v}$ and the ground-truth color $g^{j}_{u,v}$ at pixel $(u,v)$:
\begin{equation}
    e _{u,v}^{j}=\frac{1}{C^{\prime}}\sum_{c^{\prime}=1}^{C^{\prime}}{\left| r_{u,v}^{j,c^{\prime}}-g_{u,v}^{j,c^{\prime}} \right|},
\end{equation}
where $c^{\prime} \in \{1,2,\dots,C^{\prime}\}$ denotes the color channel. We then construct the loss map $\mathcal{M}^j\in \mathbb{R}^{W\times H}$ from the per-pixel errors:
\begin{equation}
\mathcal{M} ^j=\mathcal{N} \left( \{\,e_{u,v}^{j}\,\}_{u=0,v=0}^{W-1,H-1} \right) ,
\end{equation}
where $\mathcal{N} (\cdot )$ denotes a min–max normalization function.
A threshold $\tau$ is then applied to $\mathcal{M}^j$ to identify pixel $p_h$ with high reconstruction error, forming a mask:
\begin{equation}
\mathcal{M}^j_\text{mask} = \mathbb{I}(\mathcal{M}^j > \tau),
\end{equation}
where pixels $P$ with $\mathcal{M}^j_\text{mask}(u,v) = 1$ indicate regions of poor reconstruction quality.

Next, we need to find the Gaussian primitives associated with these high-error pixels. For each 3D Gaussian primitive $\mathcal{G}_i$, we project it onto the 2D image space to obtain its 2D footprint $\Omega_i$. We then use an indicator function $\mathbb{I}\big(\mathcal{M}^j_\text{mask}(p) = 1\big)$ to determine whether a pixel has high error. We compute an importance score $s^{i}_d$ for each Gaussian primitive, which accumulates the number of high-error pixels contained in the 2D footprint across all sampled views and then averages the accumulated value:
\begin{equation}
\label{eq:s+}
s_{d}^{i}=\frac{1}{K}\sum_{j=1}^K{\sum_p^{\varOmega _i}{\mathbb{I} \left( \mathcal{M} _{mask}^{j}\left( p \right) =1 \right)}},
\end{equation}
where a higher $s^{i}_d$ indicates that the Gaussian consistently lies in high-error regions across multiple views, thus suggesting it as a candidate for densification. A Gaussian primitive $\mathcal{G}_i$ is selected for densification only when its importance score $s^{i}_d$ exceeds a threshold $\tau_d$, ensuring that new Gaussians focus on under-reconstructed regions across views. Notably, we can efficiently determine the number of high-error pixels within the 2D footprint directly from the forward pass of the \texttt{render}.

\subsection{Multi-view Consistent Pruning}
\label{sec:method:pruning}
The vanilla 3DGS~\cite{kerbl3Dgaussians} removes Gaussians with low opacity or overly large scale, but cannot effectively address redundancy. Recent pruning strategies~\cite{ali2024trimming,fan2024lightgaussian,fang2024mini,salman2024elmgs,girish2024eagles} similarly fail to eliminate redundancy and can even significantly degrade rendering quality. In all cases, they do not determine Gaussian redundancy based on multi-view consistency. As illustrated in Fig.~\ref{fig:pipeline:c}, Speedy-Splat~\cite{hanson2025speedy} computes the pruning score by accumulating Gaussian-associated Hessian approximations across all training views. Hence, it leads to degraded rendering quality due to its indirect use of multi-view consistency, as visualized on the right of Fig.~\ref{fig:pipeline:c}. To remove truly redundant Gaussians, we propose a new, simple pruning strategy \VCP{}  based on multi-view consistency, as illustrated in Fig.~\ref{fig:pipeline:a}. Similar to \VCD{}, it evaluates the score according to each Gaussian’s impact on multi-view reconstruction quality. As shown on the right of Fig.~\ref{fig:pipeline:a}, VCP removes a significant number of redundant Gaussians while preserving rendering quality. We then detail how \VCP{} is implemented.

Specifically, for each view $v^j \in V$, we compute the photometric loss between the rendered image $r^j$ and the corresponding ground-truth image $g^j$: 
\begin{equation}
    E_{\mathrm{photo}}^{j} = (1-\lambda) L_1^{j} + \lambda (1-L_{\mathrm{SSIM}}^{j}),
\end{equation}
where $L_1^{j}$ and $L_{\mathrm{SSIM}}^{j}$ denote the mean absolute error and the structural similarity loss over the entire image, respectively. Since the photometric loss provides a reliable indicator of reconstruction fidelity, we incorporate it with~\cref{eq:s+} to derive the pruning score for each Gaussian primitive $\mathcal{G}_i$: 
\begin{equation}
    \label{eq:s-}
    s_{p}^{i}=\mathcal{N} \left( \sum_{j=1}^K{\left( \sum_p^{\varOmega _i}{\mathbb{I} \left( \mathcal{M} _{mask}^{j}\left( p \right) =1 \right)} \right) \cdot E_{\mathrm{photo}}^{j}} \right).
\end{equation}
Here, $s_p^{i}$ can be interpreted as a quantitative measure of the contribution of the Gaussian primitive $\mathcal{G}_i$ to the degradation of the overall rendering quality. A Gaussian primitive $\mathcal{G}_i$ is selected for pruning if its pruning score $s^{i}_p$ exceeds a predefined threshold $\tau_p$, indicating that it has relatively low contribution to rendering quality across multiple views.

\input{tables_and_figures/imags_FB}

\input{tables_and_figures/figure3_performance2}

\subsection{Compact Box}
\label{sec:method:focusedbox}
During the \texttt{preprocessing} stage of rasterization, the vanilla 3DGS \cite{kerbl3Dgaussians} uses the 3-sigma rule to obtain 2D ellipses, generating many Gaussian–tile pairs that introduce computational redundancy and reduce rendering efficiency. Speedy-Splat~\cite{hanson2025speedy} partially addresses this with precise tile intersection, yet we observe that some 2D Gaussians still have a negligible impact on pixels in certain tiles. As illustrated in Fig.~\ref{fig:fb}, to further reduce unnecessary pairs, we introduce a compact box (CB), which builds upon and extends Speedy-Splat~\cite{hanson2025speedy}’s precise tile-intersection strategy by pruning Gaussian–tile pairs with minimal contribution based on the Mahalanobis distance from the Gaussian center. This further accelerates rendering while maintaining quality. Details are provided in \cref{sec:cb} of the supplementary material.

\input{tables_and_figures/main_results}

\subsection{Optimization}
\label{sec:method:opt}
Same as the vanilla 3DGS \cite{kerbl3Dgaussians}, we optimize the learnable parameters with respect to the L1 loss over rendered pixel colors, combined with the SSIM term~\cite{SSIM} $\mathcal{L}_{\text{SSIM}}$. The total supervision is defined as:
\begin{equation}
\mathcal{L} = (1-\lambda)\mathcal{L}_1 +  \lambda(1-\mathcal{L}_{\text{SSIM}}).
\label{eq:loss}
\end{equation}

%% file: tables_and_figures/image2_framework.tex
\begin{figure*}[t]
    \centering
    \captionsetup{singlelinecheck=false}
    \includegraphics[width=\linewidth]{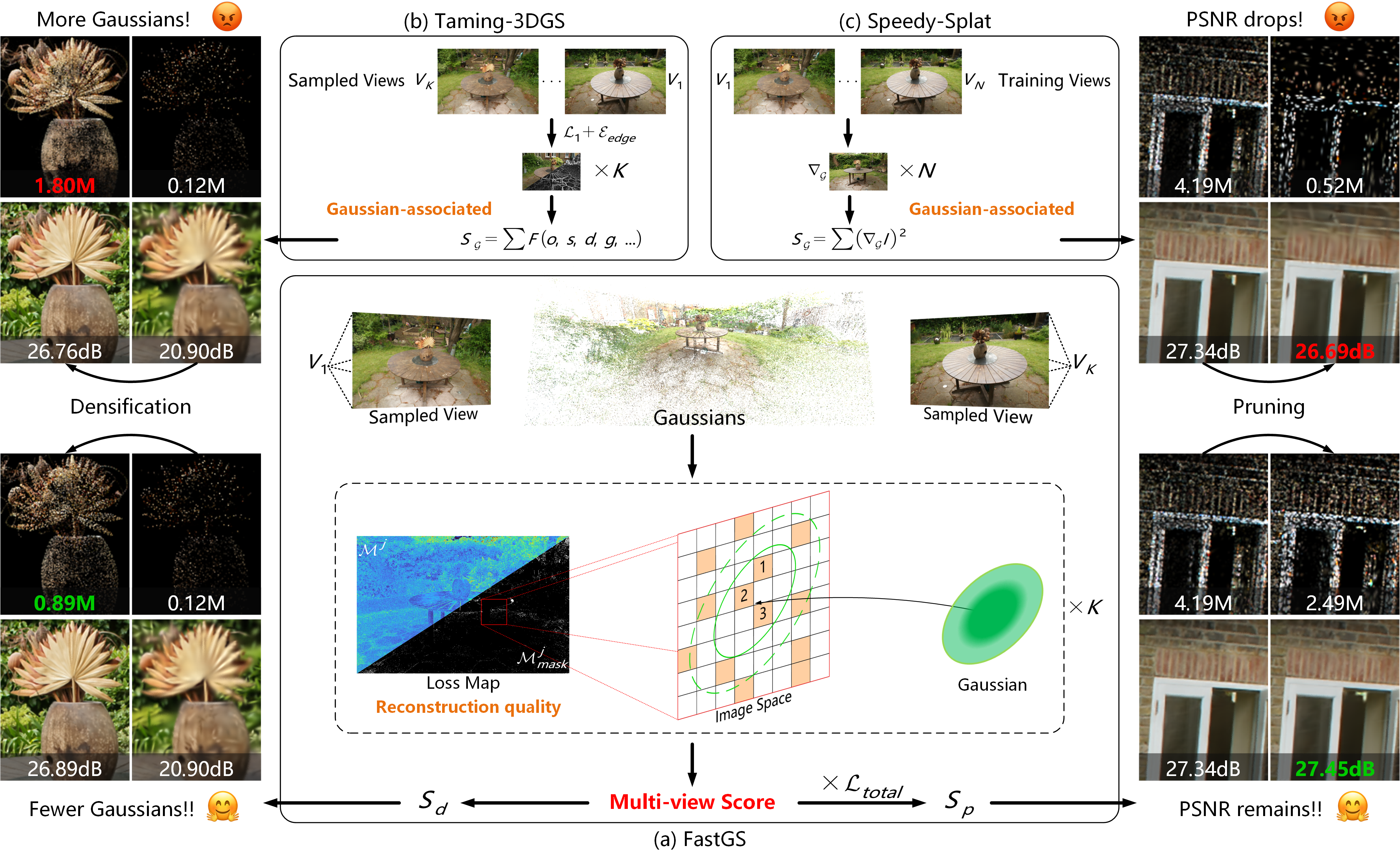}
    \caption{
        \textbf{The pipeline of \OURS{}.} (a) We redesign the ADC of the vanilla 3DGS \cite{kerbl3Dgaussians} based on multi-view consistency. To accurately assess the importance of each Gaussian, we sample training views and generate the corresponding per-pixel L1 loss maps. For each sampled view, a multi-view score is computed for each Gaussian by counting the number of high-error pixels within its 2D footprint, which is subsequently used to guide Gaussian densification and pruning. (b) Taming-3DGS~\cite{mallick2024taming} primarily computes the importance score based on Gaussian-associated properties across sampled views. (c) Speedy-Splat~\cite{hanson2025speedy} computes the Gaussian score by accumulating Gaussian-associated Hessian approximations across all training views. We visualize the densification results from 0.5K to 15K iterations without pruning on the far left, and pruning results on the far right using Speedy-Splat~\cite{hanson2025speedy}’s pruning strategy and VCP on vanilla 3DGS~\cite{kerbl3Dgaussians}. 
    }
    \label{fig:pipeline}
    \phantomsubcaption\label{fig:pipeline:a}
    \phantomsubcaption\label{fig:pipeline:b}
    \phantomsubcaption\label{fig:pipeline:c}
\end{figure*}


%% file: tables_and_figures/imags_FB.tex
\begin{figure}[t]
    \centering
    \captionsetup{singlelinecheck=false}
    \includegraphics[width=\linewidth]{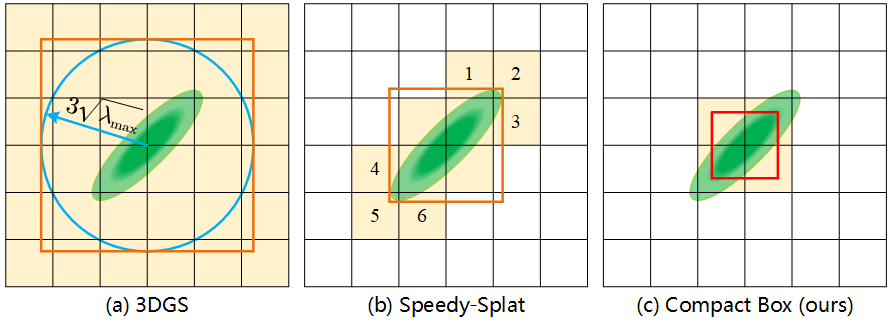}
    \caption{
       \textbf{Compact box}. Compared with vanilla 3DGS~\cite{kerbl3Dgaussians} and Speedy-Splat~\cite{hanson2025speedy}, incorporating CB leads to a reduced number of Gaussian-tile pairs.
    }
    \label{fig:fb}
    \phantomsubcaption\label{fig:fb:a}
    \phantomsubcaption\label{fig:fb:b}
    \phantomsubcaption\label{fig:fb:c}
\end{figure}

%% file: tables_and_figures/figure3_performance2.tex
\begin{figure*}
    \centering
    \captionsetup{singlelinecheck=false}
    \includegraphics[width=\linewidth]{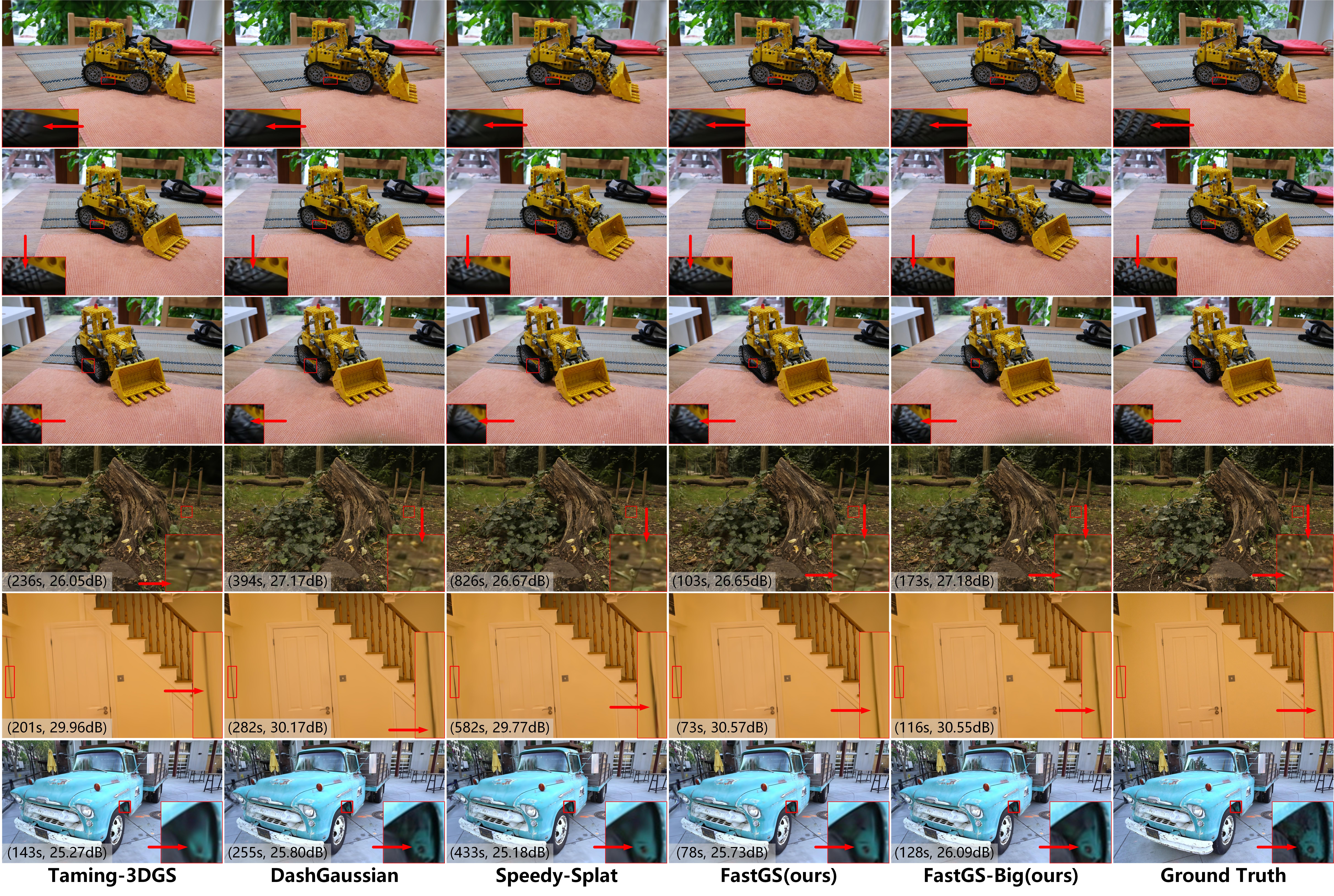}
    \vspace{-4mm}
    \caption{
        \textbf{Qualitative results of \cref{tab:main-result_}.} We present qualitative results on the \textit{kitchen} and \textit{stump} scenes from Mip-NeRF 360~\cite{barron2022mip}, the \textit{playroom} scene from Deep Blending~\cite{deng2022depth}, and the \textit{truck} scene from Tanks \& Temples~\cite{knapitsch2017tanks}. Notably, the rendered results of the \textit{kitchen} scene under multiple viewpoints demonstrate that \textbf{our method achieves more consistent details across views}.
    }
    \label{fig:per_scene}
    \vspace{-2mm}
\end{figure*}

%% file: tables_and_figures/main_results.tex
\begin{table*}[t!]
    \centering
    \footnotesize
    \setlength\tabcolsep{0.7pt}
    \captionsetup{singlelinecheck=false}
    \caption{
        \textbf{Quantitative comparisons with existing 3DGS fast optimization methods.} With FastGS, the training of 3DGS can be completed in around \textbf{100 seconds}, while achieving comparable rendering quality to the other methods. 
        Best results are marked as  \colorbox{lightred}{\bf best score}, \colorbox{lightorange}{second best score}, and \colorbox{lightyellow}{third best score}. Time is reported in minutes.
    }
    \begin{tabular} {l | rrrrrr | rrrrrr | rrrrrr}
        \toprule
        \multirow{2.5}{*}{Method} 
        & \multicolumn{6}{c}{Mip-NeRF 360~\cite{barron2022mip}}  & \multicolumn{6}{c}{Deep Blending~\cite{hedman2018deep}} & \multicolumn{6}{c}{Tanks \& Temples~\cite{knapitsch2017tanks}} \\
        \cmidrule(l{2pt}r{2pt}){2-7} \cmidrule(l{2pt}r{2pt}){8-13} \cmidrule(l{2pt}r{2pt}){14-19}
         & Time$\downarrow$ & PSNR$\uparrow$ & SSIM$\uparrow$ & LPIPS$\downarrow$ & $\mathrm{N_{GS}}\downarrow$  & FPS$\uparrow$
        & Time$\downarrow$ & PSNR$\uparrow$ & SSIM$\uparrow$ & LPIPS$\downarrow$ & $\mathrm{N_{GS}}\downarrow$  & FPS$\uparrow$
        & Time$\downarrow$ & PSNR$\uparrow$ & SSIM$\uparrow$ & LPIPS$\downarrow$ & $\mathrm{N_{GS}}\downarrow$  & FPS$\uparrow$\\
        \midrule
        
        3DGS~\cite{kerbl3Dgaussians}
            & 20.93 & 27.53 & 0.812 & 0.221 & 2.63M & 146
            & 19.77 & 29.71 & \cellcolor{tabthird}0.903 & \cellcolor{tabfirst} \bf 0.241 & 2.46M & 158 
            & 11.34 & 23.71 & \cellcolor{tabthird}0.850 & \cellcolor{tabfirst} \bf 0.170 & 1.57M & 195\\
        Mini-Splatting~\cite{fang2024mini}
            & 17.69 & 27.32 & \cellcolor{tabfirst} \bf 0.821 & \cellcolor{tabsecond}0.217 & \cellcolor{tabthird}0.53M & \cellcolor{tabsecond}567
            & 13.35 & \cellcolor{tabthird}29.99 & \cellcolor{tabfirst} \bf 0.907 & \cellcolor{tabthird}0.244 & 0.56M & \cellcolor{tabthird}624 
            & 9.06 & 23.46 & 0.844 & 0.181 & \cellcolor{tabthird}0.30M & \cellcolor{tabfirst} \bf 756 \\
        Speedy-splat~\cite{hanson2025speedy}
            & 13.38 & 26.91 & 0.781 & 0.295 & \cellcolor{tabfirst} \bf 0.30M & \cellcolor{tabthird}552
            & 10.75 & 29.42 & 0.898 & 0.272 & \cellcolor{tabsecond}0.25M & \cellcolor{tabsecond}664 
            & 6.32 & 23.38 & 0.816 & 0.242 & \cellcolor{tabfirst} \bf 0.18M & \cellcolor{tabsecond}691 \\
        Taming-3DGS~\cite{mallick2024taming}
            & \cellcolor{tabthird}5.36 & 27.48 & 0.794 & 0.261 & 0.68M & 221 
            & \cellcolor{tabthird}3.06 & 29.50 & 0.894 & 0.278 & \cellcolor{tabthird}0.29M & 352
            & \cellcolor{tabthird}2.71 & 23.89 & 0.833 & 0.214 & 0.32M & 379 \\
        DashGaussian~\cite{chen2025dashgaussian}
            & 6.35 & \cellcolor{tabsecond}27.73 & \cellcolor{tabthird}0.817 & \cellcolor{tabthird}0.218 & 2.40M & 155
            & 4.16 & 29.65 & \cellcolor{tabsecond}0.906 & 0.246 & 1.94M & 208 
            & 4.28 & \cellcolor{tabthird}24.00 & \cellcolor{tabsecond}0.853 & \cellcolor{tabthird}0.178 & 1.21M & 240 \\
        \midrule
        \OURS (Ours) 
            & \cellcolor{tabfirst} \bf 1.93 & \cellcolor{tabthird}27.56 & 0.797 & 0.261 & \cellcolor{tabsecond}0.40M & \cellcolor{tabfirst} \bf 579 
            & \cellcolor{tabfirst} \bf 1.28 & \cellcolor{tabsecond}30.03 & 0.901 & 0.270 & \cellcolor{tabfirst} \bf 0.22M & \cellcolor{tabfirst} \bf 714
            & \cellcolor{tabfirst} \bf 1.32 & \cellcolor{tabsecond}24.15 & 0.839 & 0.210 & \cellcolor{tabsecond}0.24M & \cellcolor{tabthird}655 \\
         \OURS-Big (Ours) 
            & \cellcolor{tabsecond}3.58 & \cellcolor{tabfirst} \bf 27.93 & \cellcolor{tabsecond}0.820 & \cellcolor{tabfirst} \bf 0.216 & 1.15M & 469 
            & \cellcolor{tabsecond}2.00 & \cellcolor{tabfirst} \bf 30.12 & \cellcolor{tabfirst} \bf 0.907 & \cellcolor{tabsecond}0.243 & 0.65M & 607 
            & \cellcolor{tabsecond}2.03 & \cellcolor{tabfirst} \bf 24.39 & \cellcolor{tabfirst} \bf 0.855 & \cellcolor{tabsecond}0.175 & 0.54M & 569 \\
        \bottomrule
    \end{tabular}
    \label{tab:main-result_}
\end{table*}

%% file: sec/5_experiment.tex
\section{Experiments}
\label{sec:exp}
\subsection{Experimental Setup}
\label{sec:setup}
\noindent\textbf{Datasets.}
Same as vanilla 3DGS~\cite{kerbl3Dgaussians}, we conduct experiments on three real-world datasets: Mip-NeRF 360~\cite{barron2022mip}, Deep-Blending~\cite{hedman2018deep}, and Tanks \& Temples~\cite{knapitsch2017tanks}. Moreover, we evaluate dynamic scene reconstruction on D-NeRF~\cite{pumarola2021d}, NeRF-DS~\cite{yan2023nerf}, and Neu3D~\cite{li2022neural} datasets. Tanks \& Temples~\cite{knapitsch2017tanks}, LLFF~\cite{mildenhall2019local}, BungeeNeRF~\cite{xiangli2022bungeenerf}, and Replica~\cite{replica19arxiv} are respectively used for surface reconstruction, sparse-view reconstruction, large-scale reconstruction, and SLAM.

\input{tables_and_figures/image_task}

\noindent\textbf{Metrics.}
To evaluate the performance, we report commonly used metrics for novel view rendering quality, including PSNR, SSIM~\cite{wang2004image}, and LPIPS~\cite{zhang2018unreasonable}. In addition, training efficiency and model compactness are assessed by reporting the total training time (in minutes), the final number of Gaussians, and the rendering speed (FPS). 

\input{tables_and_figures/dynamic_scene}

\noindent\textbf{Implementation Details.}
All methods, including ours and the other compared approaches, are trained for 30K iterations using the Adam~\cite{kingma2014adam} optimizer. For our approach, we set $K=10$ and $\lambda=0.2$ in all experiments. In the base setting, densification is performed every 500 iterations until the 15,000th iteration, and pruning is executed every 500 iterations before 15K and every 3,000 iterations afterwards. To ensure fairness, all experiments are conducted with an NVIDIA RTX 4090 GPU, and all comparison methods are implemented using their official code. The default configuration of \OURS{} builds upon 3DGS-accel~\cite{kerbl3Dgaussians,mallick2024taming}, detailed in \cref{sec:exp:ablation}, and incorporates the proposed \VCD{}, \VCP{}, and CB. To achieve extreme training acceleration, the rendering quality of our method is not the highest. Therefore, we provide a variant, \textbf{\OURS{}-Big}, which achieves both the highest rendering quality and the fastest training speed, where densification is executed once every 100 iterations. Further details are in \cref{sec:more_details:imp} of the supplementary material.

\subsection{Comparison with Fast Optimization Methods}
\label{sec:analysis}
\noindent\textbf{Baselines.}
We report comparisons with the SOTA fast optimization methods, including DashGaussian~\cite{chen2025dashgaussian}, Mini-Splatting~\cite{fang2024mini}, Speedy-Splat~\cite{hanson2025speedy}, and Taming-3DGS~\cite{mallick2024taming}, together with the vanilla 3DGS~\cite{kerbl3Dgaussians} for reference. These methods represent complementary approaches to accelerating training from different perspectives.

\input{tables_and_figures/table_sparse}

\noindent\textbf{Quantitative Results.}
As shown in \cref{tab:main-result_} and \cref{fig:per_scene}, \OURS{} achieves the fastest training with comparable rendering quality. A scene can be trained in around 100 seconds, and the fastest case takes only 77 seconds. 
Taming-3DGS~\cite{mallick2024taming} applies weak multi-view consistency constraints, resulting in excessive Gaussians and slower training. Similarly, the pruning strategy of Speedy-Splat~\cite{hanson2025speedy} leads to a significant drop in rendering quality. 
The current SOTA, DashGaussian~\cite{chen2025dashgaussian}, achieves high rendering quality. However, its scene optimization still retains several million Gaussians, which limits the training speed. In contrast, our variant \OURS{}-Big surpasses DashGaussian~\cite{chen2025dashgaussian} by more than 0.2 dB in rendering quality, reduces training time by 43.6\%, and cuts the number of Gaussians by more than half. These results demonstrate the superiority of our multi-view consistent densification and pruning strategies.

\subsection{Generality of \OURS{}}
\label{sec:exp:backbone_enhance}
\noindent\textbf{Baselines.} Deformable-3DGS~\cite{yang2024deformable}, PGSR~\cite{chen2024pgsr}, DropGaussian~\cite{park2025dropgaussian}, OctreeGS~\cite{ren2024octree}, and Photo-SLAM~\cite{hhuang2024photoslam} are selected as the backbones for dynamic scene reconstruction, surface reconstruction, sparse-view reconstruction, large-scale reconstruction, and SLAM, respectively.

\noindent\textbf{Enhancing Various Tasks.} We further test several SOTA methods combined with our framework across these tasks. As shown in \cref{tab:dynamic_scene}, \cref{tab:sparse-view-main}, \cref{tab:geometry_recon},  and \cref{tab:large_scale_main}, our method improves the training speed of all baselines by 2-6× while preserving rendering quality. We visualize rendered results in \cref{fig:task}, there is no degradation in rendering quality across multiple tasks. This improvement demonstrates the strong generality of our approach. We argue that this benefits from the multi-view consistency, which is fundamental to various reconstruction tasks. More details and results are provided in \cref{sec:more_details:enhance_task} of the supplementary material.

\noindent\textbf{Enhancing Backbone.}
Our framework is simple, which can be easily applied to other 3DGS backbones with different representation primitives~\cite{lu2024scaffold}, or additional filters~\cite{yu2024mip}. As shown in \cref{tab:enhance-bb}, our method achieves 3–8$\times$ faster training while maintaining the same rendering quality. 

\input{tables_and_figures/geometry_recon}
\input{tables_and_figures/table_large_scene}

\subsection{Ablation Study}
\label{sec:exp:ablation}
We adopt 3DGS-accel~\cite{kerbl3Dgaussians,mallick2024taming} as our baseline, which preserves the vanilla 3DGS~\cite{kerbl3Dgaussians} pipeline and integrates per-splat parallel backpropagation and accelerated SH optimization from Taming-3DGS~\cite{mallick2024taming}, along with optimizer scheduling. We then systematically evaluate the contribution of each proposed module based on this baseline. By adding each component individually, we analyze its impact on both reconstruction quality and training efficiency.

\noindent\textbf{Multi-View Consistent Densification.}
We first evaluate the effect of the densification strategy VCD. As shown in ~\cref{tab:ablation}, VCD achieves over 2× faster training without any loss in reconstruction quality. This is because our Gaussian addition is guided by stricter multi-view consistency, which prevents the addition of redundant Gaussians. ~\cref{tab:ablation} further validates this by showing that with VCD, the number of Gaussians is reduced by 80\%. This ablation study demonstrates the effectiveness of VCD for accelerating training.

\noindent\textbf{Multi-View Consistent Pruning.}
Next, we evaluate the effectiveness of VCP. As shown in ~\cref{tab:ablation}, adding VCP shortens the training time by 25\% and reduces the number of Gaussians by 26\%, without sacrificing rendering quality. This is because our method effectively removes redundant Gaussians while preserving those critical for scene reconstruction. The strict multi-view consistency evaluation for each deleted Gaussian ensures this effectiveness, demonstrating that VCP is highly effective.

\noindent\textbf{Compact Box.}
Finally, we evaluate the effectiveness of compact box. As shown in ~\cref{tab:ablation}, adding CB shortens the training time by 14\% while achieving comparable rendering quality. This demonstrates that CB can accelerate training without degrading reconstruction quality.

\input{tables_and_figures/table_enhance_bb}
\input{tables_and_figures/table3_ablation}

\subsection{Discussions and Limitations}
\label{sec:exp:dl}
Our method performs optimally within training from sparse point clouds. However, it faces challenges when applied to the post-training of the popular feed-forward 3DGS. Since the output Gaussians from these methods are very dense, our approach struggles to prune a massive number of points effectively within just a few thousand iterations while maintaining rendering quality, making it difficult to achieve extreme acceleration. In our tests, even a short post-training of 3K iterations still requires approximately 20 seconds.

%% file: tables_and_figures/image_task.tex
\begin{figure*}[t]
    \centering
    \captionsetup{singlelinecheck=false}
    \includegraphics[width=\linewidth]{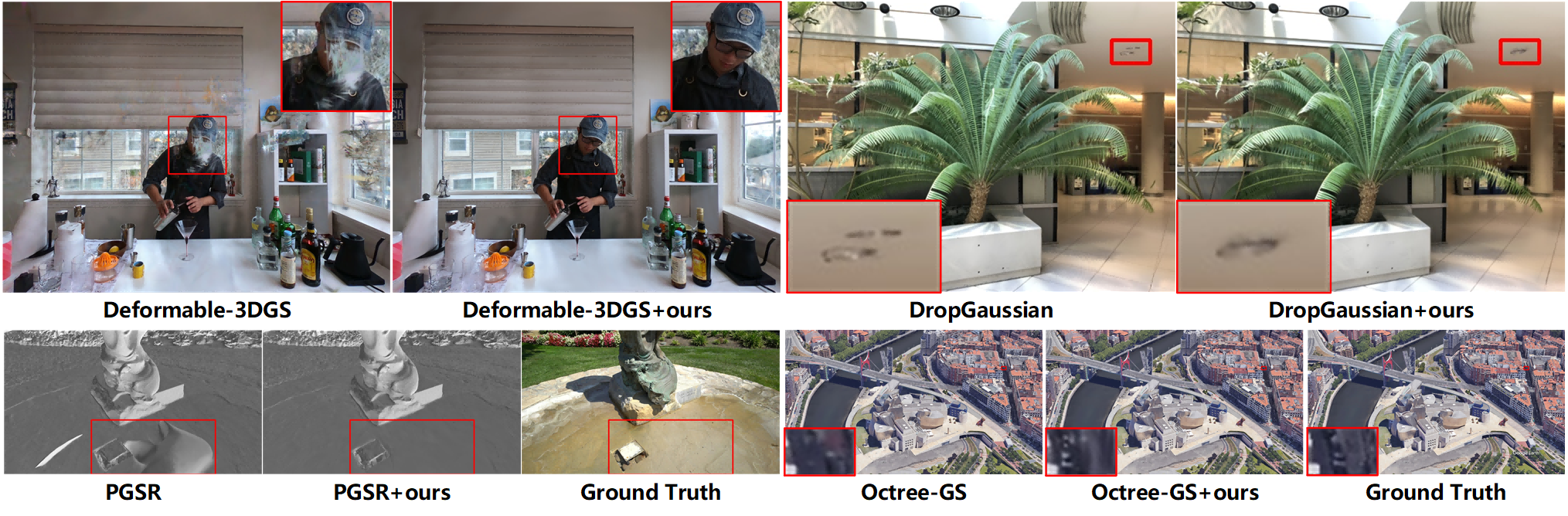}
    \caption{
        \textbf{Visualization results on four representative tasks.} We present the rendered results of the \textit{coffee-martini}, \textit{fern}, \textit{Ignatius}, and \textit{bilbao} scenes from the Neu3D~\cite{li2022neural},  LLFF~\cite{mildenhall2019local}, Tanks \& Temples~\cite{knapitsch2017tanks}, and BungeeNeRF~\cite{xiangli2022bungeenerf} datasets, respectively. 
    }
    \label{fig:task}
    \vspace{-4mm}
\end{figure*}


%% file: tables_and_figures/dynamic_scene.tex
\begin{table}[t]
    \centering
    \footnotesize
    \setlength\tabcolsep{0.1pt}
    \captionsetup{singlelinecheck=false}
    \caption{
        \textbf{Quantitative results of dynamic scene reconstruction.} Our method achieves an average 2.84$\times$ training speed-up.
    }
    \begin{tabular}{l | l | rrrrrr}
        \toprule
        Dataset & Method & Time$\downarrow$ & PSNR$\uparrow$ & SSIM$\uparrow$ & LPIPS$\downarrow$ & $\mathrm{N_{GS}}\downarrow$ & FPS$\uparrow$ \\
        \midrule
        NeRF- & Deformable-3DGS~\cite{yang2024deformable}
                  & 7.86 & 23.83 & 0.851 & 0.180 & 0.13M & 103 \\
        DS~\cite{yan2023nerf}& +Ours
                  & \textbf{2.75} & \textbf{23.90} & \textbf{0.854} & \textbf{0.176} & \textbf{0.03M} & \textbf{484} \\
        \midrule
        Neu3D & Deformable-3DGS~\cite{yang2024deformable}
                  & 26.16 & 26.74 & 0.888 & 0.168 & 0.21M & 69 \\
        ~\cite{li2022neural} & +Ours
                  & \textbf{9.29} & \textbf{29.29} & \textbf{0.908} & \textbf{0.146} & \textbf{0.09M} & \textbf{161} \\
        \bottomrule
    \end{tabular}%
    \label{tab:dynamic_scene}
    \vspace{-2mm}
\end{table}

%% file: tables_and_figures/table_sparse.tex
\begin{table}[t]
    \centering
    \footnotesize
    \setlength\tabcolsep{3pt}
    \captionsetup{singlelinecheck=false}
    \caption{
        \textbf{Quantitative results of sparse-view reconstruction.}  Our method achieves an average 2.56× training speed-up.
    }
    \begin{tabular} {l | rrrrrr}
        \toprule
        \multirow{2.5}{*}{Method} 
        & \multicolumn{6}{c}{LLFF~\cite{mildenhall2019local} (9-view)} \\
        \cmidrule(l{2pt}r{2pt}){2-7}
        & Time$\downarrow$ & PSNR$\uparrow$ & SSIM$\uparrow$ & LPIPS$\downarrow$ & $\mathrm{N_{GS}}\downarrow$  & FPS$\uparrow$\\
        \midrule
        DropGaussian~\cite{park2025dropgaussian} 
                          & 1.41 & 26.13 & \textbf{0.874} & \textbf{0.089} & 0.42M & 154   \\
        +Ours                         
                          & \textbf{0.55} & \textbf{26.14} & 0.873 & 0.095 & \textbf{0.21M} & \textbf{189}   \\
        \bottomrule
    \end{tabular}
    \label{tab:sparse-view-main}
     \vspace{-2mm}
\end{table}

%% file: tables_and_figures/geometry_recon.tex
\begin{table}[t]
    \centering
    \footnotesize
    \setlength\tabcolsep{0.5pt}
    \captionsetup{singlelinecheck=false}
    \caption{
        \textbf{Quantitative results of surface reconstruction.} 
        Our method achieves a 2-6$\times$ training speed-up.
    }
    \begin{tabular}{l | l | ccccccc}
        \toprule
        Dataset & Method & Time$\downarrow$ & PSNR$\uparrow$ & SSIM$\uparrow$ & LPIPS$\downarrow$ & $\mathrm{N_{GS}}\downarrow$ & FPS$\uparrow$ & F1$\uparrow$\\
        \midrule
        Tanks \&  & PGSR~\cite{chen2024pgsr}
                  & 32.28 & 24.20 & \textbf{0.857} & \textbf{0.149} & 1.56M & 87 & \textbf{0.57}  \\
        Temples~\cite{knapitsch2017tanks}  & +Ours
                  & \textbf{15.74} & \textbf{24.37} & 0.845 & 0.190 & \textbf{0.40M} & \textbf{210} & 0.55  \\
        \midrule
        Mip-NeRF  & PGSR~\cite{chen2024pgsr}
                  & 74.70 & 27.22 & \textbf{0.832} & \textbf{0.183} & 3.79M & 45 & - \\
        360~\cite{barron2022mip}    & +Ours
          & \textbf{11.68} & \textbf{27.23} & 0.813 & 0.240 & \textbf{0.49M} & \textbf{202} & -\\
        \bottomrule
    \end{tabular}%
    \label{tab:geometry_recon}
\end{table}

%% file: tables_and_figures/table_large_scene.tex
\begin{table}[t]
    \centering
    \footnotesize
    \setlength\tabcolsep{3.5pt}
    \captionsetup{singlelinecheck=false}
    \caption{ \textbf{Quantitative results of large-scale reconstruction.} Our method achieves an average 2.19$\times$ training speed-up.
    }
    \begin{tabular} {l | rrrrrr }
        \toprule
        \multirow{2.5}{*}{Method} 
        & \multicolumn{6}{c}{BungeeNeRF~\cite{xiangli2022bungeenerf}}   \\
        \cmidrule(l{2pt}r{2pt}){2-7} 
         & Time$\downarrow$ & PSNR$\uparrow$ & SSIM$\uparrow$ & LPIPS$\downarrow$ & $\mathrm{N_{GS}}\downarrow$  & FPS$\uparrow$\\
        \midrule
        Octree-GS~\cite{ren2024octree}
                          & 21.18 & \textbf{28.04} & \textbf{0.917} & \textbf{0.093} & 0.99M & 141  \\
        +Ours
                          & \textbf{9.68} & \textbf{28.04} & 0.910 & 0.102 & \textbf{0.74M} & \textbf{162}  \\
        \bottomrule
    \end{tabular}
    \label{tab:large_scale_main}
    \vspace{-1mm}
\end{table}


%% file: tables_and_figures/table_enhance_bb.tex
\begin{table}[t]
    \centering
    \footnotesize
    \setlength\tabcolsep{2.5pt}
    \captionsetup{singlelinecheck=false}
    \caption{
        \textbf{Quantitative results of accelerating various backbones}. Our method achieves a 3-8× training speed-up.
    }
    \begin{tabular} {l | rrrrrr }
        \toprule
        \multirow{2.5}{*}{Method} 
        & \multicolumn{6}{c}{Mip-NeRF 360~\cite{barron2022mip}}   \\
        \cmidrule(l{2pt}r{2pt}){2-7} 
         & Time$\downarrow$ & PSNR$\uparrow$ & SSIM$\uparrow$ & LPIPS$\downarrow$ & $\mathrm{N_{GS}}\downarrow$  & FPS$\uparrow$ \\
        \midrule
        Mip-Splatting~\cite{yu2024mip}
                          & 26.20 & 27.89 & \textbf{0.837} & \textbf{0.176} & 3.98M & 224         \\
        +Ours
                          & \textbf{2.98} & \textbf{27.95} & 0.828 & 0.208 & \textbf{0.83M} & \textbf{606}       \\
        \midrule
        Scaffold-GS~\cite{lu2024scaffold}
                          & 18.37 & \textbf{27.70} & \textbf{0.812} & 0.226 & 0.57M & 194       \\
        +Ours
                          & \textbf{5.06} & 27.68 & 0.809 & \textbf{0.220} & \textbf{0.30M} & \textbf{281}   \\
        \bottomrule
    \end{tabular} 
    \label{tab:enhance-bb}
\end{table}

%% file: tables_and_figures/table3_ablation.tex
\label{sec:exp:ablation}
\begin{table}[t]
    \centering
    \captionsetup{singlelinecheck=false}
    \caption{
        \textbf{Ablation studies over the proposed methods of \OURS{}.} 
        Experiments are performed on the Mip-NeRF 360 dataset~\cite{barron2022mip} with 3DGS-accel~\cite{kerbl3Dgaussians,mallick2024taming} as the baseline. 
    }
    \resizebox{\linewidth}{!}{
    
        \begin{tabular} {l | rrrrrr}
            \toprule
            Method                    & Time$\downarrow$ & PSNR$\uparrow$ & SSIM$\uparrow$ & LPIPS$\downarrow$ & $\mathrm{N_{GS}}\downarrow$  & FPS$\uparrow$ \\
            \midrule
            3DGS-accel               & 7.10 & 27.46 & 0.810 & 0.226 & 2.64M & 182 \\
            +VCD.           & 3.53 & 27.69 & 0.798 & 0.259 & 0.53M & 222 \\
            +VCP.                 & 5.32 & \textbf{27.70} & \textbf{0.812} & 0.228 & 1.96M & 285 \\
            +CB.                     & 6.13 & 27.44 & 0.810 & \textbf{0.223} & 2.78M & 303 \\
            \midrule
            Full                      & \textbf{1.93} & 27.56 & 0.797 & 0.261 & \textbf{0.40M} & \textbf{579} \\
            \bottomrule
        \end{tabular}
    }
    \label{tab:ablation}
    \vspace{-1mm}
\end{table}
    

%% file: sec/6_conclusion.tex
\vspace{0mm}
\section{Conclusion}
\label{sec:conclusion}
This paper presents a novel, simple, and general 3DGS acceleration framework \OURS{}. We propose multi-view consistent densification and pruning strategies that prevent redundant Gaussians. Extensive experiments demonstrate the effectiveness of our view-consistency design. Our method achieves the fastest training speed among all SOTA methods while maintaining comparable rendering quality. The results also demonstrate the strong generality of \OURS{}, greatly reducing training time across various tasks.

%% file: sec/X_suppl.tex
\clearpage
\setcounter{page}{1}
\maketitlesupplementary
\setcounter{section}{6}
\setcounter{table}{7}
\setcounter{figure}{6}
\section{Overview}
\label{sec:overview}
The supplementary material provides the following contents: \cref{sec:cb} gives a detailed description of the proposed compact box. \cref{sec:more_details} presents additional experimental details, where \cref{sec:more_details:imp} provides implementation details, and \cref{sec:more_details:enhance_task,sec:more_details:enhance} describe the execution details of integrating \OURS into different tasks and backbones. \cref{sec:compute_overhead} reports the computational overhead, \cref{sec:hyperpara_ablation} includes additional ablations, and \cref{sec:scene_wise_results} provides scene-wise results.

\section{Details of Compact Box}
\label{sec:cb}
During the \texttt{preprocessing} stage of rasterization, vanilla 3DGS~\cite{kerbl3Dgaussians} employs the 3-sigma rule to coarsely obtain effective 2D ellipses, which results in a large number of Gaussian–tile pairs as shown in Fig.~\ref{fig:sup_fb:a}, introducing computational redundancy and significantly reducing rendering efficiency. To address this issue, Speedy-Splat~\cite{hanson2025speedy} proposes a precise tile-intersection method to reduce the number of Gaussian–tile pairs. However, according to our observations, there remains room for further improvement, as some 2D Gaussians still have a negligible impact on the pixels within the numbered tiles in Fig.~\ref{fig:sup_fb:b}. Therefore, we propose Compact Box, which builds upon and refines Speedy-Splat’s precise tile-intersection strategy to further reduce the effective 2D Gaussian region and eliminate unnecessary Gaussian–tile pairs, as illustrated in Fig.~\ref{fig:sup_fb:c}.

Formally, during the $\alpha$-blending process, the alpha value of the $i$-th Gaussian at pixel $p$ is defined as:
\begin{equation}
\alpha_i(p) = \sigma_i \cdot \exp\Big(-\tfrac{1}{2}(p-\boldsymbol{\mu}_{i_{2D}})\boldsymbol{\Sigma}_{i_{2D}}^{-1}(p-\boldsymbol{\mu}_{i_{2D}})^{\top} \Big).
\end{equation}
This equation implies that $\alpha_i(p)$ decays exponentially with the Mahalanobis distance:
\begin{equation}
A(p) = (p-\boldsymbol{\mu}_{i_{2D}})\boldsymbol{\Sigma}_{i_{2D}}^{-1}(p-\boldsymbol{\mu}_{i_{2D}})^{\top}.
\end{equation}
Intuitively, pixels closer to the Gaussian center contribute more, while those farther away have negligible influence. Based on this observation, a reasonable criterion can be established: Gaussian–tile pairs corresponding to pixels with large Mahalanobis distances can be safely discarded, as their effect on the final rendering is minimal.

To prune Gaussian–tile pairs whose contributions are negligible, we define a threshold for the Mahalanobis distance as:
\begin{equation}
\label{eq:cb_beta}
\left( p-\mu _{i_{2D}} \right) \varSigma _{_{i_{2D}}}^{-1}\left( p-\mu _{i_{2D}} \right) ^T=\beta \left( 2\ln \small{\frac{\sigma _i}{\tau _{\alpha}}} \right),
\end{equation}
where $\beta$ is a scaling factor. By adjusting $\beta$, the effective support region of each 2D Gaussian can be flexibly controlled. A smaller $\beta$ yields a tighter ellipse around the mean $\mu _{i_{2D}}$, thereby reducing the spatial extent of $(p-\mu _{i_{2D}})$ and limiting the number of pixels influenced by the Gaussian $\mathcal{G}_j$. This selective suppression of marginal Gaussian contributions effectively reduces redundant Gaussian–tile pairs and accelerates rasterization. 

In implementation, ~\cref{eq:cb_beta} is integrated into Speedy-Splat~\cite{hanson2025speedy}’s snugbox, where the parameter $\beta$ further reduces the 2D Gaussian footprint and shrinks its intersection region with tiles, thus forming our CB. We then obtain the Gaussian–tile pairs using Speedy-Splat~\cite{hanson2025speedy}’s accutile method. The comparison between snugbox~\cite{hanson2025speedy} and our CB can be found in \cref{tab:cs}. We sincerely appreciate the excellent work of Speedy-Splat~\cite{hanson2025speedy}.

\input{tables_and_figures/sup_image_FB}

\section{More Details}
\label{sec:more_details}
\subsection{Implementation Details}
\label{sec:more_details:imp}
Our \OURS{} integrate VCD, VCP, and CB into 3DGS-accel~\cite{kerbl3Dgaussians,mallick2024taming} and adopt the widely used absolute gradients~\cite{ye2024absgs} to ensure more accurate densification, ultimately achieving an average training time of around 100 seconds per scene. The core of our method lies in strictly controlling the number of Gaussians throughout training via VCD and VCP, maintaining it at a very low level and thereby enabling significant acceleration, as shown in Fig.~\hyperref[fig:gaussian_count]{\textcolor[rgb]{0.21,0.49,0.74}{2}}.

\input{tables_and_figures/few_shots}
In practice, our VCD and VCP introduce a multi-view consistency constraint that strictly regulates Gaussian densification and pruning, preventing the creation of redundant Gaussians.  For densification, newly added Gaussians are required not only to satisfy the gradient-based criteria but also to have an importance score greater than $\tau_d$, which is set to 5 in our experiments. We follow the vanilla 3DGS~\cite{kerbl3Dgaussians} gradient for cloning, while adopting the absolute gradient from AbsGS~\cite{ye2024absgs} for splitting. For pruning, before 15k iterations, we follow 3DGS~\cite{kerbl3Dgaussians} but retain VCP by sampling half of the candidate Gaussians according to their pruning scores. After 15k iterations, pruning is performed every 3k iterations by removing Gaussians with opacity below 0.1 or pruning score exceeding 0.9, ensuring multi-view consistency.

Our baseline, 3DGS-accel~\cite{kerbl3Dgaussians,mallick2024taming}, preserves the vanilla 3DGS~\cite{kerbl3Dgaussians} pipeline while integrating the per-splat parallel backpropagation and accelerated SH optimization from Taming-3DGS~\cite{mallick2024taming}, along with an optimizer schedule that updates the optimizer every 32 iterations from 15,000 to 20,000 iterations and every 64 iterations thereafter. This schedule is inspired by the SH optimization strategy in Taming-3DGS~\cite{mallick2024taming}. We do not consider it a conceptual contribution of our method, so instead of discussing it in the main paper, we incorporate it directly into the baseline configuration. As shown in \cref{tab:optimizer}, this scheduling strategy provides acceleration comparable to sparse Adam~\cite{mallick2024taming} while fully preserving rendering quality. 

We sincerely thank Taming-3DGS~\cite{mallick2024taming} for providing a strong baseline, upon which our work builds. By further integrating the proposed VCD, VCP, and CB components, together with the absolute gradients from AbsGS~\cite{ye2024absgs}, our method achieves significant acceleration. This improvement largely stems from the strict control of Gaussian count enforced by VCD and VCP, ensuring that the number of Gaussians remains as low as possible throughout training. Other modules contribute only marginally to this aspect, which we further analyze in the ablation study.

\subsection{Generalizing \OURS to Other Tasks}
\label{sec:more_details:enhance_task}
\noindent\textbf{Dynamic Scene Reconstruction:}
Deformable-3DGS~\cite{yang2024deformable} adopts the same ADC strategy as vanilla 3DGS~\cite{kerbl3Dgaussians}, while additionally predicting per-Gaussian deformation parameters. This design remains fully compatible with our framework. Building on our accelerated 3DGS backbone, we employ VCD and VCP to precisely regulate densification and pruning, ensuring that the number of Gaussians remains low throughout training. CB is further integrated to speed up rendering, and the optimizer is Adam~\cite{kingma2014adam}.

\noindent\textbf{Surface Reconstruction:} PGSR~\cite{chen2024pgsr} uses an ADC strategy similar to vanilla 3DGS~\cite{kerbl3Dgaussians}. Based on our accelerated 3DGS backbone, we replace ADC with VCD and VCP to strictly control Gaussian growth and elimination. CB is integrated into the rasterization stage to further improve efficiency. Adam~\cite{kingma2014adam} is used as the optimizer.

\noindent\textbf{Sparse-view Reconstruction:} DropGaussian~\cite{park2025dropgaussian} differs from vanilla 3DGS~\cite{kerbl3Dgaussians} in that it randomly sets the opacity of a subset of Gaussians to zero during rendering, eliminating their contribution. Based on our accelerated 3DGS backbone, we apply VCD and VCP to precisely control densification and pruning, keeping the number of Gaussians low throughout training. CB is incorporated to accelerate rendering, and Adam~\cite{kingma2014adam} is used as the optimizer. Additional experimental results are presented in \cref{tab:few-shots-llff}.

\noindent\textbf{Large-scale Reconstruction:} Octree-GS~\cite{ren2024octree} uses an anchor-based parameterization where each anchor generates multiple Gaussians. Based on our accelerated 3DGS backbone, we apply VCD to constrain anchor expansion, requiring the importance score of associated Gaussians to exceed 5. VCP is not applied since pruning is performed at the anchor level. CB is integrated into rasterization. The optimizer is Adam~\cite{kingma2014adam}.

\noindent\textbf{SLAM:} Photo-SLAM~\cite{hhuang2024photoslam} follows vanilla 3DGS~\cite{kerbl3Dgaussians}’s ADC strategy. Based on our accelerated 3DGS backbone, we integrate VCD and VCP for effective densification and pruning, and incorporate CB to speed up rendering. Adam~\cite{kingma2014adam} is used as the optimizer. Results are presented in \cref{tab:slam}.
\input{tables_and_figures/table_slam}
\input{tables_and_figures/table2_enhanceBackbone}
\input{tables_and_figures/table_GPU}
\subsection{Equipping \OURS to Backbones}
\label{sec:more_details:enhance}
\noindent\textbf{Mip-Splatting~\cite{yu2024mip}:} Mip-Splatting~\cite{yu2024mip} introduces a filtering mechanism for anti-aliasing while following an ADC strategy similar to vanilla 3DGS~\cite{kerbl3Dgaussians}. Based on our accelerated 3DGS backbone, we replace its ADC pipeline with VCD and VCP, which precisely control Gaussian densification and pruning to keep the number of Gaussians low throughout training. CB is also integrated into the rasterization stage to further accelerate rendering. Adam~\cite{kingma2014adam} is used as the optimizer. Additional experimental results are presented in \cref{tab:second-result}.

\noindent\textbf{Scaffold-GS~\cite{lu2024scaffold}:} Scaffold-GS~\cite{lu2024scaffold} adopts an anchor-based Gaussian representation. On our accelerated 3DGS backbone, we apply VCD to control densification and maintain a low number of Gaussians. VCP is not applicable because pruning is performed at the anchor level. CB is integrated into the rasterization stage to further accelerate rendering. Adam~\cite{kingma2014adam} is used as the optimizer. Additional experimental results are presented in \cref{tab:second-result}.

\section{Computational Overhead}
\label{sec:compute_overhead}
We report computational resource consumption in \cref{tab:gpu}. As shown, our method requires relatively low GPU memory, making it suitable for devices with limited resources.

\input{tables_and_figures/table_ablation1} 

\section{Additional Ablation}
\label{sec:hyperpara_ablation}
In this section, we perform more comprehensive ablations based on 3DGS-accel~\cite{kerbl3Dgaussians,mallick2024taming} to further demonstrate that the proposed multi-view consistency-based densification and pruning strategies, VCD and VCP, contribute most significantly to the overall acceleration.

\noindent \textbf{Component-wise Ablation.} We examine the effects of VCD, VCP, CB, and the absolute gradients from AbsGS~\cite{ye2024absgs} in \cref{tab:ablation1}. 
The ablation results indicate that neither absolute gradients nor CB effectively reduce the number of Gaussians, and their contribution to acceleration is limited. In contrast, \textbf{our proposed VCD and VCP achieve significantly greater speed-up, as they strictly control the Gaussian count, keeping it low throughout the entire training process, as shown in Fig.~\hyperref[fig:gaussian_count]{\textcolor[rgb]{0.21,0.49,0.74}{2}}.
}

\noindent \textbf{VCD Threshold $\tau_d$.} We study the effect of the densification threshold $\tau_d$ in VCD on Tanks \& Temples~\cite{knapitsch2017tanks}, as shown in \cref{tab:ablation_taud}. A smaller $\tau_d$ allows more Gaussians to be densified, leading to slightly higher rendering quality but at the cost of increased training time and Gaussian count. Conversely, a larger $\tau_d$ reduces the number of Gaussians, accelerating training while slightly degrading quality. Our default choice of $\tau_d = 5$ achieves a balanced trade-off between efficiency and rendering fidelity.
\input{tables_and_figures/table_ablation_taud}

\noindent \textbf{Number of sampled views $K$.} We study the effect of the number of sampled views on both training efficiency and rendering quality on the Mip-NeRF 360~\cite{barron2022mip} dataset. As shown in \cref{tab:ablation_k}, using too few views slightly degrades quality, while sampling more views increases training time with minimal improvement. Our default choice of $K=10$ achieves a good balance. 
\input{tables_and_figures/table_ablation_K}

\input{tables_and_figures/table_optimizer}
\input{tables_and_figures/table_cs}

\section{Scene-wise Results}
We present the quantitative results in \cref{tab:scenewise-mip360}, \cref{tab:scenewise-db}, and \cref{tab:scenewise-tanks}, and provide the qualitative comparisons in Fig.~\ref{fig:sup_figure_1}, Fig.~\ref{fig:sup_figure_21} and Fig.~\ref{fig:sup_figure_22}.
\label{sec:scene_wise_results}
\input{tables_and_figures/table_scenewise_mip360}
\input{tables_and_figures/table_scenewise_db}
\input{tables_and_figures/table_scenewise_tanks}
\input{tables_and_figures/sup_figure_1}
\input{tables_and_figures/sup_figure_2}
\input{tables_and_figures/sup_figure_22}

%% file: tables_and_figures/sup_image_FB.tex
\begin{figure}[t]
    \centering
    \captionsetup{singlelinecheck=false}
    \includegraphics[width=\linewidth]{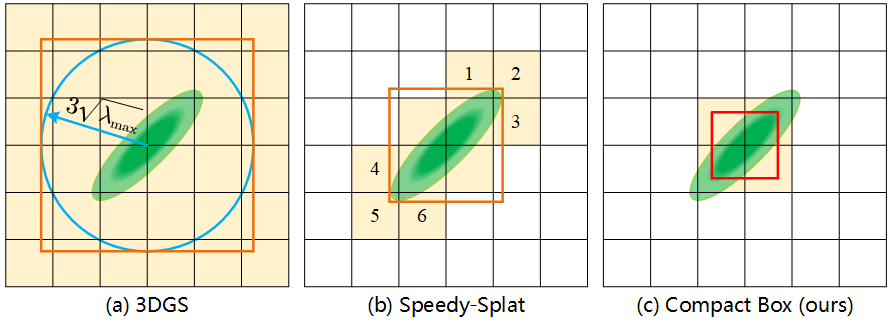}
    \caption{
       \textbf{Compact box}. Compared with vanilla 3DGS~\cite{kerbl3Dgaussians} and Speedy-Splat~\cite{hanson2025speedy}, incorporating CB leads to a reduced number of Gaussian-tile pairs.
    }
    \label{fig:sup_fb}
    \phantomsubcaption\label{fig:sup_fb:a}
    \phantomsubcaption\label{fig:sup_fb:b}
    \phantomsubcaption\label{fig:sup_fb:c}
\end{figure}

%% file: tables_and_figures/few_shots.tex
\begin{table*}[t]
    \centering
    \footnotesize
    \captionsetup{singlelinecheck=false}
    \caption{
        \textbf{Quantitative results of sparse-view reconstruction.} We present the results under the 3-view and 6-view settings.
    }
    \resizebox{\textwidth}{!}{
    \begin{tabular} {l | rrrrrr | rrrrrr}
        \toprule
        \multirow{2.5}{*}{Method} 
        & \multicolumn{6}{c}{LLFF~\cite{mildenhall2019local} 3-view}  & \multicolumn{6}{c}{LLFF~\cite{mildenhall2019local} 6-view}  \\
        \cmidrule(l{2pt}r{2pt}){2-7} \cmidrule(l{2pt}r{2pt}){8-13} 
         & Time$\downarrow$ & PSNR$\uparrow$ & SSIM$\uparrow$ & LPIPS$\downarrow$ & $\mathrm{N_{GS}}\downarrow$  & FPS$\uparrow$
        & Time$\downarrow$ & PSNR$\uparrow$ & SSIM$\uparrow$ & LPIPS$\downarrow$ & $\mathrm{N_{GS}}\downarrow$  & FPS$\uparrow$\\
        \midrule
        DropGaussian~\cite{park2025dropgaussian}
                          & 0.73 & 20.43 & 0.707 & \textbf{0.202} & 0.08M & 183       
                          & 0.88 & 24.67 & \textbf{0.836} & \textbf{0.116} & 0.18M & 174   \\
        +Ours
                          & \textbf{0.30} & \textbf{20.58} & \textbf{0.708} & 0.217 & \textbf{0.03M} & \textbf{206}       
                          & \textbf{0.37} & \textbf{24.68} & 0.834 & 0.131 & \textbf{0.09M} & \textbf{199}  \\
        \bottomrule
    \end{tabular}
    }
    \label{tab:few-shots-llff}
\end{table*}

%% file: tables_and_figures/table_slam.tex
\begin{table}[t]
    \centering
    \footnotesize
    \setlength\tabcolsep{0.7pt}
    \captionsetup{singlelinecheck=false}
    \caption{ \textbf{Quantitative results of SLAM.} Our method achieves an average 2.70$\times$ training speed-up.
    }
    \resizebox{\columnwidth}{!}{
    \begin{tabular} {l | rrrrrr }
        \toprule
        \multirow{2.5}{*}{Method} 
        & \multicolumn{6}{c}{Replica RGB-D}   \\
        \cmidrule(l{2pt}r{2pt}){2-7} 
         & Time$\downarrow$ & PSNR$\uparrow$ & SSIM$\uparrow$ & LPIPS$\downarrow$ & $\mathrm{N_{GS}}\downarrow$  & FPS$\uparrow$\\
        \midrule
        Photo-SLAM~\cite{hhuang2024photoslam}
                          & 5.03 & \textbf{37.01} & \textbf{0.961} & \textbf{0.026} & 0.33M & 744   \\
        +Ours
                          & \textbf{1.86} & \textbf{37.01} & 0.957 & 0.042 & \textbf{0.11M} & \textbf{2700}  \\
        \bottomrule
    \end{tabular}
    }
    \label{tab:slam}
    \vspace{-1mm}
\end{table}

%% file: tables_and_figures/table2_enhanceBackbone.tex
\begin{table*}[t]
    \centering
    \footnotesize
    \captionsetup{singlelinecheck=false}
    \caption{
    \textbf{Quantitative results of accelerating various backbones}. 
    }
    \resizebox{\textwidth}{!}{
    \begin{tabular} {l | rrrrrr | rrrrrr}
        \toprule
        \multirow{2.5}{*}{Method} 
        & \multicolumn{6}{c}{Deep Blending~\cite{hedman2018deep}} & \multicolumn{6}{c}{Tanks \& Temples~\cite{knapitsch2017tanks}} \\
        \cmidrule(l{2pt}r{2pt}){2-7} \cmidrule(l{2pt}r{2pt}){8-13} 
        & Time$\downarrow$ & PSNR$\uparrow$ & SSIM$\uparrow$ & LPIPS$\downarrow$ & $\mathrm{N_{GS}}\downarrow$  & FPS$\uparrow$
        & Time$\downarrow$ & PSNR$\uparrow$ & SSIM$\uparrow$ & LPIPS$\downarrow$ & $\mathrm{N_{GS}}\downarrow$  & FPS$\uparrow$\\
        \midrule
        Mip-Splatting~\cite{yu2024mip}     
                          & 23.94 & 29.35 & \textbf{0.899} & \textbf{0.241} & 3.48M & 219   
                          & 14.57 & 23.77 & \textbf{0.856} & \textbf{0.158} & 2.36M & 300   \\
        +Ours
                          & \textbf{1.74} & \textbf{29.68} & \textbf{0.899} & 0.274 & \textbf{0.20M} & \textbf{698}   
                          & \textbf{2.01} & \textbf{24.18} & 0.843 & 0.200 & \textbf{0.36M} & \textbf{729}   \\
        \midrule
        Scaffold-GS~\cite{lu2024scaffold}    
                          & 13.31 & \textbf{30.09} & \textbf{0.905} & \textbf{0.256} & 0.18M & 307   
                          & 9.83 & 24.09 & \textbf{0.851} & \textbf{0.175} & 0.26M & 261   \\
        +Ours
                          & \textbf{2.82} & 30.00 & 0.900 & 0.267 & \textbf{0.08M} & \textbf{423}   
                          & \textbf{3.77} & \textbf{24.15} & 0.849 & 0.180 & \textbf{0.14M} & \textbf{332}   \\
        \bottomrule
    \end{tabular}
   }
    \label{tab:second-result}
\end{table*}

%% file: tables_and_figures/table_GPU.tex
\begin{table*}[t!]
    \centering
    \captionsetup{singlelinecheck=false}
    \caption{
        \textbf{Quantitative comparison of computational overhead.} We report the mean GPU memory usage (GB), peak GPU memory usage (GB), and storage size (MB). 
    }
    \resizebox{\textwidth}{!}{
    \begin{tabular} {l | ccc | ccc | ccc}
        \toprule
        \multirow{3}{*}{Method} 
        & \multicolumn{3}{c}{MipNeRF-360~\cite{barron2022mip}}  & \multicolumn{3}{c}{Deep Blending~\cite{hedman2018deep}} & \multicolumn{3}{c}{Tanks\&Temples~\cite{knapitsch2017tanks}} \\
        \cmidrule(l{2pt}r{2pt}){2-4} \cmidrule(l{2pt}r{2pt}){5-7} \cmidrule(l{2pt}r{2pt}){8-10}
         &  mean GPU mem.$\downarrow$& peak GPU mem.$\downarrow$&  Storage$\downarrow$ 
         &  mean GPU mem.$\downarrow$& peak GPU mem.$\downarrow$&  Storage$\downarrow$
         &  mean GPU mem.$\downarrow$& peak GPU mem.$\downarrow$&  Storage$\downarrow$\\
        \midrule
        
        3DGS~\cite{kerbl3Dgaussians}
            & 7.70 & 9.89 & 652
            & 5.97 & 8.10 & 610 
            & 3.47 & 4.73 & 389 \\
        Mini-Splatting~\cite{fang2024mini}
            & 5.21 & 7.44 & \cellcolor{tabthird}132
            & 4.16 & 6.20 & 138 
            & 2.51 & 4.63 & \cellcolor{tabthird}75 \\
        Speedy-splat~\cite{hanson2025speedy}
            & \cellcolor{tabthird}4.97 & 7.03 & \cellcolor{tabfirst} \bf 74
            & 3.82 & 5.03 & \cellcolor{tabsecond}61 
            & \cellcolor{tabthird}2.19 & \cellcolor{tabthird}2.66 & \cellcolor{tabfirst} \bf 45 \\
        Taming-3DGS~\cite{mallick2024taming}
            & \cellcolor{tabsecond}4.78 & \cellcolor{tabsecond}5.88 & 170
            & \cellcolor{tabsecond}3.39 & \cellcolor{tabsecond}4.01 & \cellcolor{tabthird}73 
            & \cellcolor{tabsecond}1.94 & \cellcolor{tabsecond}2.43 & 79 \\
        DashGaussian~\cite{chen2025dashgaussian}
            & 6.71 & 9.96 & 595
            & 4.79 & 7.75 & 482 
            & 2.81 & 4.49 & 301 \\
        \midrule
        \OURS{} (Ours) 
            & \cellcolor{tabfirst} \bf 4.58 & \cellcolor{tabfirst} \bf 5.21 & \cellcolor{tabsecond}99
            & \cellcolor{tabfirst} \bf 3.34 & \cellcolor{tabfirst} \bf 3.77 & \cellcolor{tabfirst} \bf 54 
            & \cellcolor{tabfirst} \bf 1.91 & \cellcolor{tabfirst} \bf 2.27 & \cellcolor{tabsecond}60 \\
        \OURS{}-Big (Ours) 
            & 5.37 & \cellcolor{tabthird}6.63 & 208
            & \cellcolor{tabthird}3.81 & \cellcolor{tabthird}4.65 & 114
            & 2.22 & 2.83 & 86 \\
        \bottomrule
    \end{tabular}
    }
    \label{tab:gpu}
\end{table*}

%% file: tables_and_figures/table_ablation1.tex
\label{sec:exp:ablation}
\begin{table}[t]
    \centering
    \captionsetup{singlelinecheck=false}
    \caption{
        \textbf{Ablation studies over the proposed methods.} 
        Experiments are performed on the Mip-NeRF 360 dataset~\cite{barron2022mip} with 3DGS-accel~\cite{kerbl3Dgaussians,mallick2024taming} as the baseline. 
    }
    \resizebox{\linewidth}{!}{
    
        \begin{tabular} {l | rrrrr}
            \toprule
            Method                    & Time$\downarrow$ & PSNR$\uparrow$ & SSIM$\uparrow$ & LPIPS$\downarrow$ & $\mathrm{N_{GS}}\downarrow$  \\
            \midrule
            3DGS-accel               & 7.10 & 27.46 & 0.810 & 0.226 & 2.64M  \\
            \midrule
            +Abs grad              & 6.85 & 27.60 & \textbf{0.817} & \textbf{0.216} & 2.29M  \\
            +CB.                     & 6.13 & 27.44 & 0.810 & 0.223 & 2.78M  \\
            +VCD.           & 3.53 & 27.69 & 0.798 & 0.259 & 0.53M  \\
            +VCP.                 & 5.32 & \textbf{27.70} & 0.812 & 0.228 & 1.96M  \\
            \midrule
            Full                      & \textbf{1.93} & 27.56 & 0.797 & 0.261 & \textbf{0.40M}  \\
            \bottomrule
        \end{tabular}
    }
    \label{tab:ablation1}
\end{table}
    

%% file: tables_and_figures/table_ablation_taud.tex
\label{sec:exp:ablation}
\begin{table}[t]
    \centering
    \captionsetup{singlelinecheck=false}
    \caption{
        \textbf{Ablation study on thresh $\tau_d$ on Tanks \& Temples~\cite{knapitsch2017tanks}.} 
    }
    \resizebox{\linewidth}{!}{
    
        \begin{tabular} {l | rrrrr}
            \toprule
            $\tau_d$                    & Time$\downarrow$ & PSNR$\uparrow$ & SSIM$\uparrow$ & LPIPS$\downarrow$ & $\mathrm{N_{GS}}\downarrow$   \\
            \midrule
            1                  & 1.44 & 24.17 & \textbf{0.841} & \textbf{0.205} & 0.30M\\
            2                  & 1.42 & \textbf{24.18} & \textbf{0.841} & 0.207 & 0.28M\\
            5(ours)                  & 1.32 & 24.15 & 0.839 & 0.210 & 0.24M\\
            10                 & 1.30 & 23.98 & 0.834 & 0.218 & 0.21M\\
            20                 & 1.23 & 23.84 & 0.829 & 0.226 & 0.17M\\
            50                 & 1.15 & 23.54 & 0.819 & 0.241 & 0.13M\\
            100                & \textbf{1.12} & 23.26 & 0.809 & 0.252 & \textbf{0.11M}\\
            \bottomrule
        \end{tabular}
    }
    \label{tab:ablation_taud}
\end{table}

%% file: tables_and_figures/table_ablation_K.tex
\label{sec:exp:ablation}
\begin{table}[t]
    \centering
    \captionsetup{singlelinecheck=false}
    \caption{
        \textbf{Ablation study on the number of sampled views $K$ on Mip-NeRF 360~\cite{barron2022mip}.} 
        ``all'' indicates using all training views.
    }
    \resizebox{\linewidth}{!}{
    
        \begin{tabular} {l | rrrrr}
            \toprule
            $K$                    & Time$\downarrow$ & PSNR$\uparrow$ & SSIM$\uparrow$ & LPIPS$\downarrow$ & $\mathrm{N_{GS}}\downarrow$   \\
            \midrule
            5                     & \textbf{1.91} & 27.47 & 0.795 & 0.265 & \textbf{0.36M} \\
            10(ours)              & 1.93 & \textbf{27.56} & \textbf{0.797} & \textbf{0.261} & 0.40M \\
            20                    & 2.03 & 27.55 & \textbf{0.797} & \textbf{0.261} & 0.43M \\
            50                    & 2.10 & 27.55 & \textbf{0.797} & 0.262 & 0.43M \\
            all                  & 2.29 & 27.54 & 0.796 & 0.263 & 0.42M \\
            \bottomrule
        \end{tabular}
    }
    \label{tab:ablation_k}
\end{table}

%% file: tables_and_figures/table_optimizer.tex
\label{sec:exp:ablation}
\begin{table}[t]
    \centering
    \captionsetup{singlelinecheck=false}
    \caption{
        \textbf{Comparison of different optimization strategies.} 
    }
    \resizebox{\linewidth}{!}{
        \begin{tabular} {ll | rrrrr}
            \toprule
            Method & Optimizer                    & Time$\downarrow$ & PSNR$\uparrow$ & SSIM$\uparrow$ & LPIPS$\downarrow$ & $\mathrm{N_{GS}}\downarrow$   \\
            \midrule
            \multirow{2}{*}{FastGS} 
            & Sparse Adam~\cite{mallick2024taming}             & \textbf{1.93} & 27.37 & 0.792 & 0.270 & \textbf{0.37M} \\
            & Optimizer schedule       & \textbf{1.93} & \textbf{27.56} & \textbf{0.797} & \textbf{0.261} & 0.40M  \\
            \bottomrule
        \end{tabular}
    }
    \label{tab:optimizer}
\end{table}

%% file: tables_and_figures/table_cs.tex
\begin{table}[t]
    \centering
    \captionsetup{singlelinecheck=false}
        \caption{
    \textbf{Comparison of rasterization methods.} 
    Experiments are performed on the Mip-NeRF 360 dataset~\cite{barron2022mip} with 3DGS-accel~\cite{kerbl3Dgaussians,mallick2024taming} as the baseline.
}
    \resizebox{\linewidth}{!}{
    
        \begin{tabular} {l | rrrrrr}
            \toprule
            Method                    & Time$\downarrow$ & PSNR$\uparrow$ & SSIM$\uparrow$ & LPIPS$\downarrow$ & $\mathrm{N_{GS}}\downarrow$ & FPS$\uparrow$  \\
            \midrule
            3DGS-accel               & 7.10 & \textbf{27.46} & 0.810 & 0.226 & \textbf{2.64M} & 182 \\
            \midrule
            +snugbox~\cite{hanson2025speedy}     & 6.50 & \textbf{27.46} & \textbf{0.811} & 0.224 & 2.69M & 288 \\
            +CB.                     & \textbf{6.13} & 27.44 & 0.810 & \textbf{0.223} & 2.78M  & \textbf{303} \\
            \bottomrule
        \end{tabular}
    }
    \label{tab:cs}
\end{table}

%% file: tables_and_figures/table_scenewise_mip360.tex
\begin{table*}[t]
    \centering
    \footnotesize
    \setlength\tabcolsep{0.7pt}
    \captionsetup{singlelinecheck=false}
    \caption{
        Scene-wise quantitative results over the Mip-NeRF 360 dataset~\cite{barron2022mip}. 
    }
    \begin{tabular} {l | rrrrrr | rrrrrr | rrrrrr}
        \toprule
        \multirow{2}{*}{Method} 
        & \multicolumn{6}{c}{bicycle} & \multicolumn{6}{c}{flowers} & \multicolumn{6}{c}{garden} \\
        \cmidrule(l{2pt}r{2pt}){2-7} \cmidrule(l{2pt}r{2pt}){8-13} \cmidrule(l{2pt}r{2pt}){14-19}
        & Time$\downarrow$ & PSNR$\uparrow$ & SSIM$\uparrow$ & LPIPS$\downarrow$ & $\mathrm{N_{GS}}\downarrow$ & FPS$\uparrow$ 
        & Time$\downarrow$ & PSNR$\uparrow$ & SSIM$\uparrow$ & LPIPS$\downarrow$ & $\mathrm{N_{GS}}\downarrow$ & FPS$\uparrow$  
        & Time$\downarrow$ & PSNR$\uparrow$ & SSIM$\uparrow$ & LPIPS$\downarrow$ & $\mathrm{N_{GS}}\downarrow$ & FPS$\uparrow$   \\
        \midrule
        3DGS~\cite{kerbl3Dgaussians}
                          & 27.97 & 25.14 & 0.748 & \cellcolor{tabthird}0.242 & 4.71M & 80
                          & 18.98  & 21.30 & 0.586 & \cellcolor{tabsecond}0.360 & 2.82M & 162
                          & 26.78 & 27.34 & \cellcolor{tabsecond}0.857 & \cellcolor{tabsecond}0.122 & 4.19M & 103\\
        Mini-Splatting~\cite{fang2024mini}
                          & 16.17 & \cellcolor{tabthird}25.23 & \cellcolor{tabfirst}0.764 & \cellcolor{tabsecond}0.241 & \cellcolor{tabthird}0.59M & \cellcolor{tabsecond}564
                          & 17.22 & \cellcolor{tabthird}21.43 & \cellcolor{tabfirst}0.614 & \cellcolor{tabfirst}0.341 & 0.63M & \cellcolor{tabthird}511
                          & 15.97 & 27.36 & 0.806 & 0.215 & \cellcolor{tabsecond}0.67M & \cellcolor{tabsecond}487 \\
        Speedy-Splat~\cite{hanson2025speedy}
                          & 15.87 & 24.79 & 0.704 & 0.333 & \cellcolor{tabsecond}0.58M & 460 
                          & 13.38 & 21.21 & 0.560 & 0.418 & \cellcolor{tabfirst}0.34M & \cellcolor{tabsecond}526
                          & 15.73 & 26.69 & 0.814 & 0.214 & \cellcolor{tabfirst}0.52M & \cellcolor{tabthird}474 \\
        Taming-3DGS~\cite{mallick2024taming}
                          & \cellcolor{tabthird}5.65 & 24.72 & 0.693 & 0.332 & 0.81M & 199
                          & \cellcolor{tabthird}4.97 & 21.10 & 0.552 & 0.416 & \cellcolor{tabthird}0.58M & 233
                          & 9.82 & \cellcolor{tabthird}27.42 & \cellcolor{tabthird}0.851 & 0.138 & 2.08M & 177 \\
        DashGaussian~\cite{chen2025dashgaussian}
                          & 9.93 & \cellcolor{tabfirst}25.31 & \cellcolor{tabsecond}0.763 & \cellcolor{tabfirst}0.222 & 4.70M & 105
                          & 7.05 & \cellcolor{tabfirst}21.78 & \cellcolor{tabsecond}0.604 & \cellcolor{tabfirst}0.341 & 2.82M & 158
                          & \cellcolor{tabthird}8.27 & \cellcolor{tabfirst}27.57 & \cellcolor{tabsecond}0.857 & \cellcolor{tabthird}0.131 & 3.37M & 153 \\
        \midrule
        \OURS
                          & \cellcolor{tabfirst}1.92 & 24.84 & 0.714 & 0.310 & \cellcolor{tabfirst}0.54M & \cellcolor{tabfirst}582
                          & \cellcolor{tabfirst}1.95 & 21.21 & 0.560 & \cellcolor{tabthird}0.406 & \cellcolor{tabsecond}0.49M & \cellcolor{tabfirst}555
                          & \cellcolor{tabfirst}2.47 & 27.20 & 0.836 & 0.174 & \cellcolor{tabthird}0.74M & \cellcolor{tabfirst}538 \\
        \OURS{}-Big
                          & \cellcolor{tabsecond}2.59 & \cellcolor{tabsecond}25.26 & \cellcolor{tabthird}0.755 & 0.245 & 1.55M & \cellcolor{tabthird}463
                          & \cellcolor{tabsecond}3.22 & \cellcolor{tabsecond}21.60 & \cellcolor{tabthird}0.602 & \cellcolor{tabfirst}0.341 & 1.14M & 468
                          & \cellcolor{tabsecond}6.50 & \cellcolor{tabsecond}27.56 & \cellcolor{tabfirst}0.864 & \cellcolor{tabfirst}0.110 & 2.64M & 332 \\
        \midrule
        \midrule
        \multirow{2}{*}{Method} 
        & \multicolumn{6}{c}{stump} & \multicolumn{6}{c}{treehill} & \multicolumn{6}{c}{room} \\
        \cmidrule(l{2pt}r{2pt}){2-7} \cmidrule(l{2pt}r{2pt}){8-13} \cmidrule(l{2pt}r{2pt}){14-19}
        & Time$\downarrow$ & PSNR$\uparrow$ & SSIM$\uparrow$ & LPIPS$\downarrow$ & $\mathrm{N_{GS}}\downarrow$ & FPS$\uparrow$ 
        & Time$\downarrow$ & PSNR$\uparrow$ & SSIM$\uparrow$ & LPIPS$\downarrow$ & $\mathrm{N_{GS}}\downarrow$ & FPS$\uparrow$  
        & Time$\downarrow$ & PSNR$\uparrow$ & SSIM$\uparrow$ & LPIPS$\downarrow$ & $\mathrm{N_{GS}}\downarrow$ & FPS$\uparrow$   \\
        \midrule
        3DGS~\cite{kerbl3Dgaussians}
                          & 21.77 & 26.64 & 0.768 & 0.244 & 4.05M & 130
                          & 20.33 & 22.59 & \cellcolor{tabthird}0.636 & \cellcolor{tabthird}0.347 & 3.01M & 136
                          & 18.78 & 31.71 & \cellcolor{tabthird}0.927 & \cellcolor{tabthird}0.197 & 1.25M & 164 \\
        Mini-Splatting~\cite{fang2024mini}
                          & 16.52 & \cellcolor{tabthird}26.80 & \cellcolor{tabfirst}0.839 & \cellcolor{tabfirst}0.161 & 0.67M & \cellcolor{tabsecond}521 
                          & 17.05 & 22.76 & \cellcolor{tabfirst}0.656 & \cellcolor{tabfirst}0.326 & \cellcolor{tabthird}0.63M & 487 
                          & 18.00 & 31.48 & \cellcolor{tabsecond}0.928 & \cellcolor{tabsecond}0.190 & 0.39M & 506 \\
        Speedy-Splat~\cite{hanson2025speedy}
                          & 13.77 & 26.67 & 0.765 & 0.288 & \cellcolor{tabsecond}0.46M & 480
                          & 12.90 & 22.48 & 0.590 & 0.462 & \cellcolor{tabfirst}0.32M & \cellcolor{tabsecond}548      
                          & 12.05 & 30.83 & 0.903 & 0.258 & \cellcolor{tabfirst}0.11M & \cellcolor{tabsecond}617\\
        Taming-3DGS~\cite{mallick2024taming}
                          & \cellcolor{tabthird}3.93 & 26.05 & 0.729 & 0.324 & \cellcolor{tabthird}0.48M & 280
                          & \cellcolor{tabthird}5.37 & \cellcolor{tabsecond}22.92 & 0.628 & 0.395 & 0.79M & 214
                          & \cellcolor{tabthird}3.88 & 31.64 & 0.917 & 0.227 & \cellcolor{tabthird}0.23M & 230 \\
        DashGaussian~\cite{chen2025dashgaussian}
                          & 6.57 & \cellcolor{tabsecond}27.17 & \cellcolor{tabthird}0.783 & \cellcolor{tabfirst}0.229 & 3.42M & 164
                          & 8.20 & \cellcolor{tabfirst}22.94 & \cellcolor{tabsecond}0.640 & \cellcolor{tabsecond}0.333 & 3.42M & 134
                          & 4.00 & \cellcolor{tabthird}31.81 & 0.924 & 0.205 & 1.04M & 182 \\
        \midrule
        \OURS
                          & \cellcolor{tabfirst}1.72 & 26.65 & 0.756 & 0.297 & \cellcolor{tabfirst}0.39M & \cellcolor{tabfirst}576
                          & \cellcolor{tabfirst}1.72 & \cellcolor{tabfirst}22.94 & 0.612 & 0.429 & \cellcolor{tabsecond}0.38M & \cellcolor{tabfirst}568
                          & \cellcolor{tabfirst}1.62 & \cellcolor{tabsecond}31.98 & 0.920 & 0.217 & \cellcolor{tabsecond}0.21M & \cellcolor{tabfirst}632 \\
        \OURS{}-Big
                          & \cellcolor{tabsecond}2.88 & \cellcolor{tabfirst}27.18 & \cellcolor{tabsecond}0.786 & \cellcolor{tabthird}0.240 & 1.06M & \cellcolor{tabthird}489
                          & \cellcolor{tabsecond}2.78 & \cellcolor{tabthird}22.83 & 0.632 & 0.378 & 1.01M & \cellcolor{tabthird}507
                          & \cellcolor{tabsecond}2.38 & \cellcolor{tabfirst}32.20 & \cellcolor{tabfirst}0.929 & \cellcolor{tabfirst}0.189 & 0.57M & \cellcolor{tabthird}577 \\
        \midrule
        \midrule
        \multirow{2}{*}{Method} 
        & \multicolumn{6}{c}{counter} & \multicolumn{6}{c}{kitchen} & \multicolumn{6}{c}{bonsai} \\
        \cmidrule(l{2pt}r{2pt}){2-7} \cmidrule(l{2pt}r{2pt}){8-13} \cmidrule(l{2pt}r{2pt}){14-19}
        & Time$\downarrow$ & PSNR$\uparrow$ & SSIM$\uparrow$ & LPIPS$\downarrow$ & $\mathrm{N_{GS}}\downarrow$ & FPS$\uparrow$ 
        & Time$\downarrow$ & PSNR$\uparrow$ & SSIM$\uparrow$ & LPIPS$\downarrow$ & $\mathrm{N_{GS}}\downarrow$ & FPS$\uparrow$  
        & Time$\downarrow$ & PSNR$\uparrow$ & SSIM$\uparrow$ & LPIPS$\downarrow$ & $\mathrm{N_{GS}}\downarrow$ & FPS$\uparrow$   \\
        \midrule
        3DGS~\cite{kerbl3Dgaussians}
                          & 17.58 & \cellcolor{tabthird}29.16 & \cellcolor{tabsecond}0.915 & \cellcolor{tabthird}0.183 & 1.05M & 171
                          & 21.30 & 31.54 & \cellcolor{tabsecond}0.932 & \cellcolor{tabsecond}0.116 & 1.53M & 144
                          & 14.90 & \cellcolor{tabthird}32.37 & \cellcolor{tabsecond}0.946 & \cellcolor{tabthird}0.180 & 1.07M & 224 \\
        Mini-Splatting~\cite{fang2024mini}
                          & 9.83 & 28.65 & \cellcolor{tabthird}0.911 & \cellcolor{tabsecond}0.181 & 0.41M & \cellcolor{tabthird}590
                          & 10.23 & 31.05 & \cellcolor{tabthird}0.930 & \cellcolor{tabthird}0.120 & \cellcolor{tabthird}0.44M & \cellcolor{tabfirst}614
                          & 11.02 & 31.24 & 0.943 & \cellcolor{tabsecond}0.177 & \cellcolor{tabthird}0.36M & \cellcolor{tabfirst}661 \\
        Speedy-Splat~\cite{hanson2025speedy}
                          & 12.10 & 28.22 & 0.876 & 0.259 & \cellcolor{tabfirst}0.10M & \cellcolor{tabfirst}606
                          & 13.25 & 30.09 & 0.895 & 0.195 & \cellcolor{tabfirst}0.11M & \cellcolor{tabsecond}608
                          & 11.37 & 31.16 & 0.925 & 0.228 & \cellcolor{tabfirst}0.13M & \cellcolor{tabsecond}652\\
        Taming-3DGS~\cite{mallick2024taming}
                          & 4.60 & \cellcolor{tabsecond}29.20 & 0.909 & 0.200 & \cellcolor{tabthird}0.31M & 221
                          & \cellcolor{tabsecond}3.48 & \cellcolor{tabthird}31.84 & 0.929 & 0.128 & 0.48M & 209
                          & 4.60 & \cellcolor{tabsecond}32.40 & 0.942 & 0.193 & 0.41M & 227\\
        DashGaussian~\cite{chen2025dashgaussian}
                          & \cellcolor{tabthird}3.95 & 29.11 & \cellcolor{tabthird}0.911 & 0.191 & 0.85M & 162
                          & 5.52 & 31.69 & 0.927 & 0.129 & 1.18M& 135
                          & \cellcolor{tabthird}3.95 & 32.15 & \cellcolor{tabthird}0.945 & \cellcolor{tabthird}0.180 & 0.82M & 193 \\
        \midrule
        \OURS
                          & \cellcolor{tabfirst}1.83 & 29.15 & 0.907 & 0.204 & \cellcolor{tabsecond}0.21M & \cellcolor{tabsecond}596
                          & \cellcolor{tabfirst}2.42 & \cellcolor{tabsecond}31.87 & 0.929 & 0.127 & \cellcolor{tabsecond}0.38M& \cellcolor{tabthird}543
                          & \cellcolor{tabfirst}1.83 & 32.19 & 0.942 & 0.191 & \cellcolor{tabsecond}0.28M & \cellcolor{tabthird}622\\
        \OURS{}-Big
                          & \cellcolor{tabsecond}2.62 & \cellcolor{tabfirst}29.57 & \cellcolor{tabfirst}0.917 & \cellcolor{tabfirst}0.177 & 0.47M & 522
                          & \cellcolor{tabthird}5.15 & \cellcolor{tabfirst}32.17 & \cellcolor{tabfirst}0.938 & \cellcolor{tabfirst}0.105 & 1.18M & 395 
                          & \cellcolor{tabsecond}3.22 & \cellcolor{tabfirst}32.97 & \cellcolor{tabfirst}0.953 & \cellcolor{tabfirst}0.161 & 0.85M & 498\\
        \bottomrule
    \end{tabular}
    \label{tab:scenewise-mip360}
\end{table*}

%% file: tables_and_figures/table_scenewise_db.tex
\begin{table*}[t]
    \centering
    \footnotesize
    \captionsetup{singlelinecheck=false}
    \caption{
        Scene-wise quantitative results over the Deep Blending dataset~\cite{deng2022depth}. 
    }
    \begin{tabular} {l | rrrrrr | rrrrrr}
        \toprule
        \multirow{2}{*}{Method} 
        & \multicolumn{6}{c}{playroom} & \multicolumn{6}{c}{drjohnson} \\
        \cmidrule(l{2pt}r{2pt}){2-7} \cmidrule(l{2pt}r{2pt}){8-13}
        & Time$\downarrow$ & PSNR$\uparrow$ & SSIM$\uparrow$ & LPIPS$\downarrow$ & $\mathrm{N_{GS}}\downarrow$ & FPS$\uparrow$ 
        & Time$\downarrow$ & PSNR$\uparrow$ & SSIM$\uparrow$ & LPIPS$\downarrow$ & $\mathrm{N_{GS}}\downarrow$ & FPS$\uparrow$  
      \\
        \midrule
        3DGS~\cite{kerbl3Dgaussians}
                          & 16.75 & 30.14 & 0.904 & \cellcolor{tabthird}0.243 & 1.85M & 189 
                          & 22.78 & 29.28 & \cellcolor{tabthird}0.902 & \cellcolor{tabfirst}0.239 & 3.07M & 126
                          \\
        Mini-Splatting~\cite{fang2024mini}
                          & 12.30 & \cellcolor{tabthird}30.47 & \cellcolor{tabsecond}0.908 & \cellcolor{tabsecond}0.241 & 0.51M & 618
                          & 14.40 & \cellcolor{tabsecond}29.51 & \cellcolor{tabfirst}0.905 & \cellcolor{tabsecond}0.246 & 0.60M & \cellcolor{tabthird}629
                          \\
        Speedy-Splat~\cite{hanson2025speedy}
                          & 9.70 & 29.77 & 0.898 & 0.274 & \cellcolor{tabfirst}0.18M & \cellcolor{tabsecond}695
                          & 11.80 & 29.07 & 0.898 & 0.269 & \cellcolor{tabthird}0.31M & \cellcolor{tabsecond}633
                          \\
        Taming-3DGS~\cite{mallick2024taming}
                          & \cellcolor{tabthird}3.35 & 29.96 & 0.901 & 0.264 & \cellcolor{tabthird}0.40M & 354 
                          & \cellcolor{tabthird}2.77 & 29.04 & 0.888 & 0.292 & \cellcolor{tabfirst}0.19M & 349
                          \\
        DashGaussian~\cite{chen2025dashgaussian}
                          & 4.70 & 30.17 & \cellcolor{tabfirst}0.909 & \cellcolor{tabthird}0.243 & 2.38M & 233
                          & 3.62 & 29.13 & \cellcolor{tabsecond}0.903 & 0.250 & 1.51M & 182
                          \\
        \midrule
        \OURS
                         & \cellcolor{tabfirst}1.22 & \cellcolor{tabfirst}30.57 & \cellcolor{tabthird}0.905 & 0.266 & \cellcolor{tabsecond}0.19M & \cellcolor{tabfirst}727
                         & \cellcolor{tabfirst}1.35 & \cellcolor{tabthird}29.50 & 0.898 & 0.275 & \cellcolor{tabsecond}0.25M & \cellcolor{tabfirst}700 
                          \\
        \OURS{}-Big
                          & \cellcolor{tabsecond}1.93 & \cellcolor{tabsecond}30.55 & \cellcolor{tabfirst}0.909 & \cellcolor{tabfirst}0.239 & 0.60M & \cellcolor{tabthird}619
                          & \cellcolor{tabsecond}2.07 & \cellcolor{tabfirst}29.69 & \cellcolor{tabfirst}0.905 & \cellcolor{tabthird}0.247 & 0.70M & 595
                          \\
        \bottomrule
    \end{tabular}
    \label{tab:scenewise-db}
\end{table*}

%% file: tables_and_figures/table_scenewise_tanks.tex
\begin{table*}[t]
    \centering
    \footnotesize
    \captionsetup{singlelinecheck=false}
    \caption{
        Scene-wise quantitative results over the Tanks \& Temples dataset~\cite{knapitsch2017tanks}. 
    }
    \begin{tabular} {l | rrrrrr | rrrrrr}
        \toprule
        \multirow{2}{*}{Method} 
        & \multicolumn{6}{c}{truck} & \multicolumn{6}{c}{train} \\
        \cmidrule(l{2pt}r{2pt}){2-7} \cmidrule(l{2pt}r{2pt}){8-13}
        & Time$\downarrow$ & PSNR$\uparrow$ & SSIM$\uparrow$ & LPIPS$\downarrow$ & $\mathrm{N_{GS}}\downarrow$ & FPS$\uparrow$ 
        & Time$\downarrow$ & PSNR$\uparrow$ & SSIM$\uparrow$ & LPIPS$\downarrow$ & $\mathrm{N_{GS}}\downarrow$ & FPS$\uparrow$ \\
        \midrule
        3DGS~\cite{kerbl3Dgaussians}
                          & 12.50 & 25.39 & \cellcolor{tabsecond}0.881 & \cellcolor{tabthird}0.142 & 2.05M & 186
                          & 10.18 & 22.03 & \cellcolor{tabthird}0.818 & \cellcolor{tabfirst}0.198 & 1.09M & 203
                          \\
        Mini-Splatting~\cite{fang2024mini}
                          & 9.08 & 25.32 & \cellcolor{tabthird}0.879 & \cellcolor{tabfirst}0.139 & 0.32M & \cellcolor{tabfirst}716
                          & 9.03 & 21.60 & 0.809 & 0.223 & \cellcolor{tabthird}0.28M & \cellcolor{tabfirst}795
                          \\
        Speedy-Splat~\cite{hanson2025speedy}
                          & 7.22 & 25.18 & 0.863 & 0.192 & \cellcolor{tabsecond}0.26M & \cellcolor{tabthird}648
                          & 5.42 & 21.59 & 0.768 & 0.292 & \cellcolor{tabfirst}0.11M & \cellcolor{tabsecond}733
                          \\
        Taming-3DGS~\cite{mallick2024taming}
                          & \cellcolor{tabthird}2.38 & 25.27 & 0.865 & 0.187 & \cellcolor{tabthird}0.27M & 409
                          & \cellcolor{tabthird}3.03 & \cellcolor{tabthird}22.50 & 0.802 & 0.240 & 0.37M & 348
                          \\
        DashGaussian~\cite{chen2025dashgaussian}
                          & 4.25 & \cellcolor{tabsecond}25.80 & \cellcolor{tabfirst}0.886 & 0.150 & 1.43M & 257
                          & 4.32 & 22.19 & \cellcolor{tabsecond}0.819 & \cellcolor{tabsecond}0.206 & 1.00M & 222
                          \\
        \midrule
        \OURS
                          & \cellcolor{tabfirst}1.30 & \cellcolor{tabthird}25.73 & 0.872 & 0.178 & \cellcolor{tabfirst}0.25M & \cellcolor{tabsecond}666
                          & \cellcolor{tabfirst}1.33 & \cellcolor{tabsecond}22.57 & 0.805 & 0.242 & \cellcolor{tabsecond}0.23M & \cellcolor{tabthird}644
                          \\
        \OURS{}-Big
                          & \cellcolor{tabsecond}2.13 & \cellcolor{tabfirst}26.09 & \cellcolor{tabfirst}0.886 & \cellcolor{tabsecond}0.140 & 0.63M & 579
                          & \cellcolor{tabsecond}1.93 & \cellcolor{tabfirst}22.68 & \cellcolor{tabfirst}0.824 & \cellcolor{tabthird}0.210 & 0.46M & 558
                          \\
        \bottomrule
    \end{tabular}
    \label{tab:scenewise-tanks}
\end{table*}

%% file: tables_and_figures/sup_figure_1.tex
\begin{figure*}
    \centering
    \captionsetup{singlelinecheck=false}
    \includegraphics[width=\linewidth]{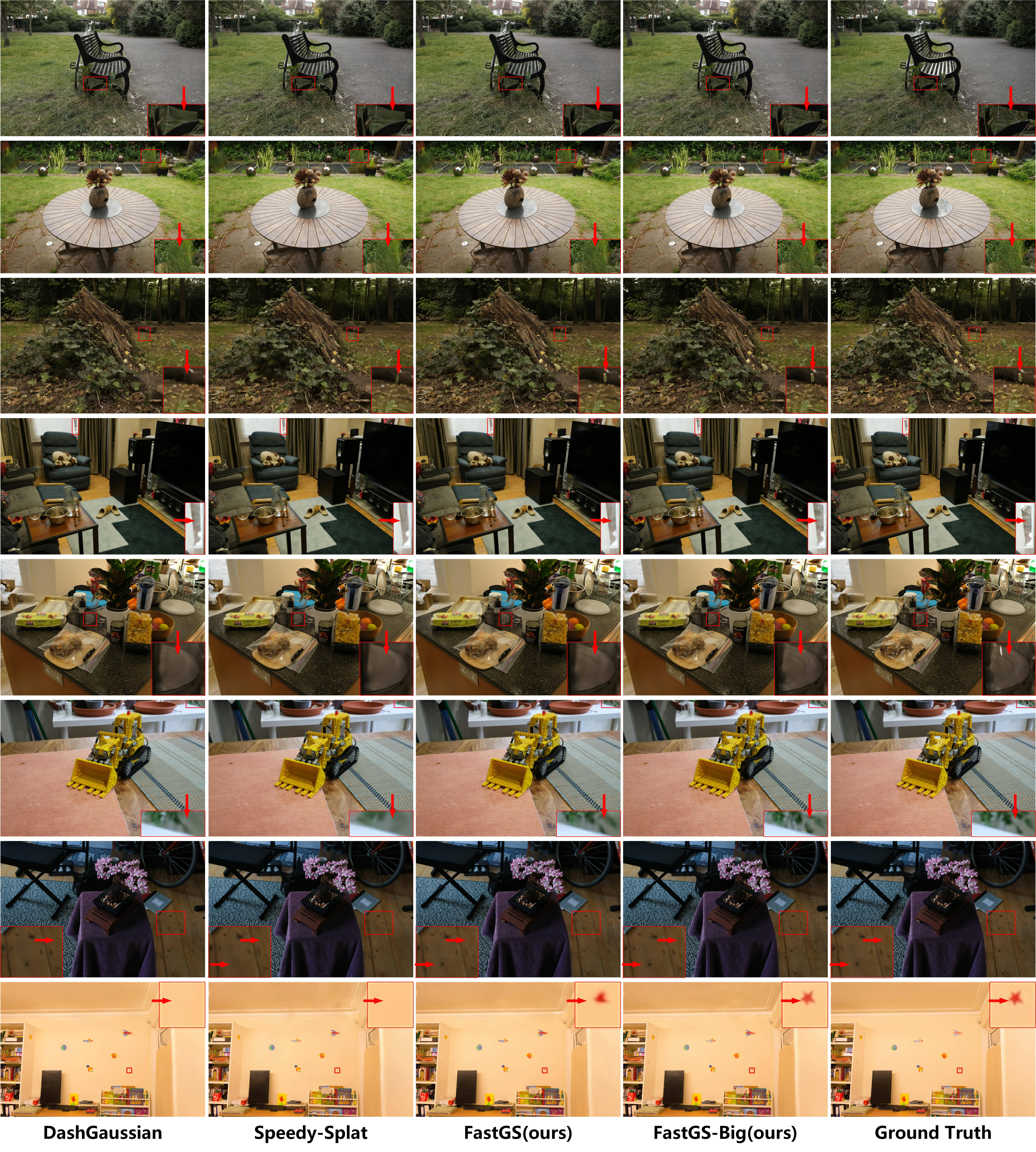}
    \caption{
        Additional visual comparisons on the \textit{bicycle}, \textit{garden}, \textit{stump}, \textit{room}, \textit{counter}, \textit{kitchen}, \textit{bonsai}, and \textit{playroom} scenes.
    }
    \label{fig:sup_figure_1}
\end{figure*}

%% file: tables_and_figures/sup_figure_2.tex
\begin{figure*}
    \centering
    \captionsetup{singlelinecheck=false}
    \includegraphics[width=\linewidth]{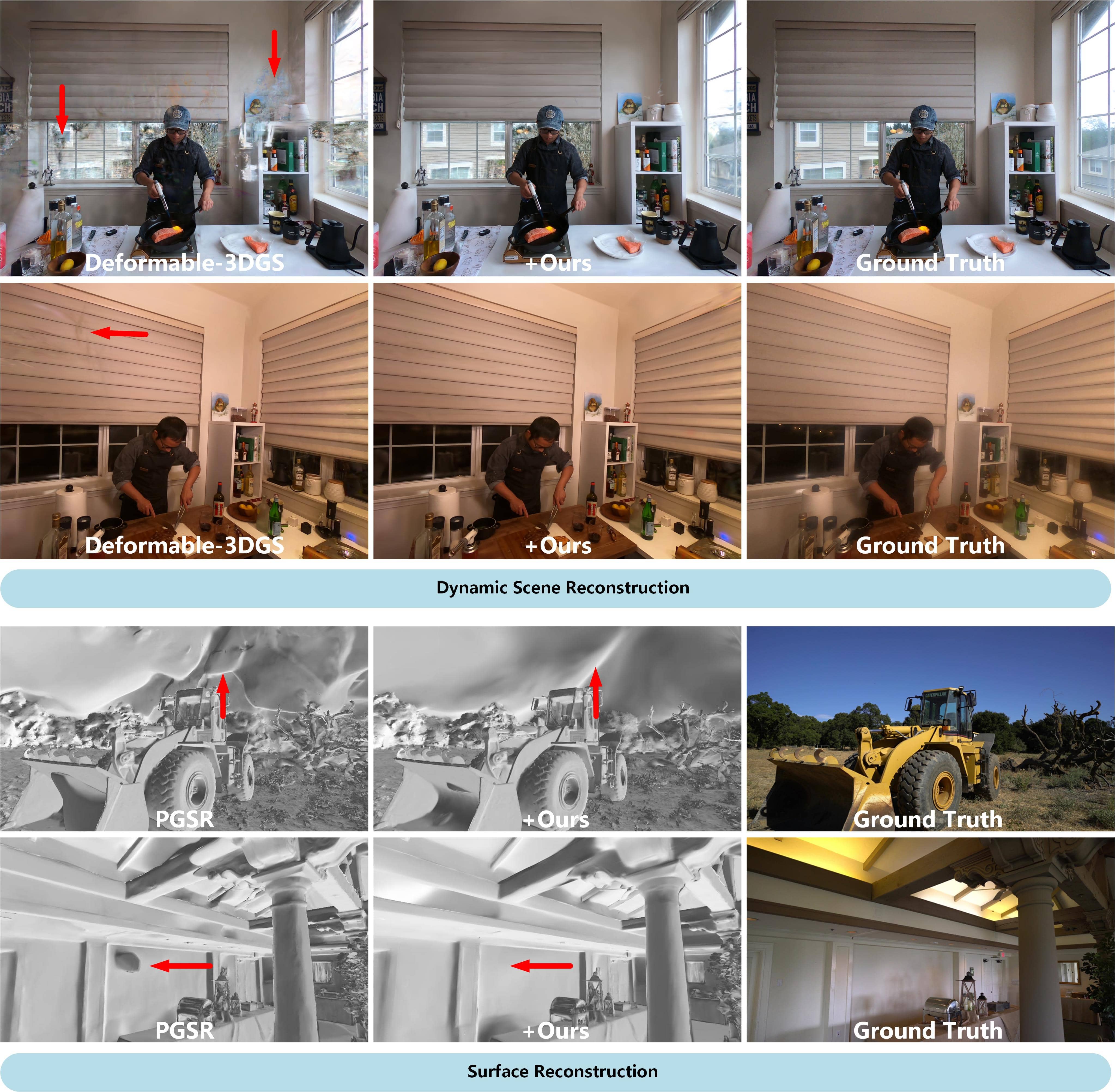}
    \caption{Additional visual comparisons on different tasks, including the \textit{flame\_salmon}, \textit{cut\_roasted\_beef}, \textit{Caterpillar}, \textit{Meetingroom} scenes.
    }
    \label{fig:sup_figure_21}
\end{figure*}

%% file: tables_and_figures/sup_figure_22.tex
\begin{figure*}
    \centering
    \captionsetup{singlelinecheck=false}
    \includegraphics[width=\linewidth]{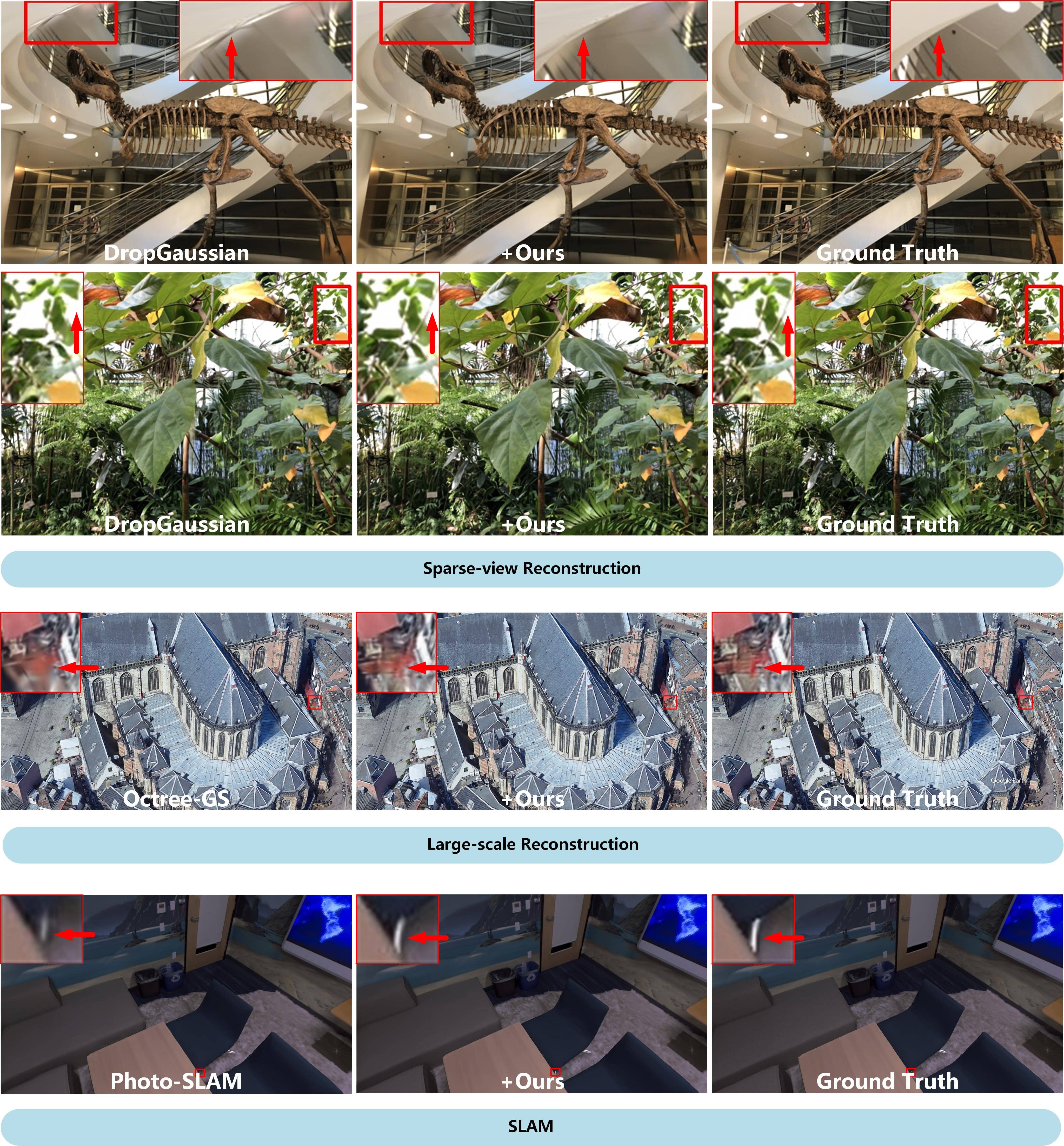}
    \caption{Additional visual comparisons on different tasks, including the \textit{trex}, \textit{leaves}, \textit{amsterdam}, and \textit{office0} scenes.
    }
    \label{fig:sup_figure_22}
\end{figure*}